\documentclass[conference]{IEEEtran}
\IEEEoverridecommandlockouts
\usepackage{cite}
\usepackage{amsmath,amssymb,amsfonts}
\usepackage{algorithmic}
\usepackage{graphicx}
\usepackage{textcomp}
\usepackage{xcolor}
\usepackage{subcaption}
\usepackage{hyperref}
\usepackage{multirow}
\usepackage{makecell}

\def\BibTeX{{\rm B\kern-.05em{\sc i\kern-.025em b}\kern-.08em
    T\kern-.1667em\lower.7ex\hbox{E}\kern-.125emX}}
\begin{document}

\title{Sparsely-gated Mixture-of-Expert Layers for CNN Interpretability
}

\author{\IEEEauthorblockN{Svetlana Pavlitska$^{1,2}$, Christian Hubschneider$^{1,2}$, Lukas Struppek$^{2}$, J.~Marius Zöllner$^{1,2}$}
\IEEEauthorblockA{
\textit{$^{1}$ FZI Research Center for Information Technology} \\
 \textit{$^{2}$ Karlsruhe Institute of Technology (KIT)}\\
Karlsruhe, Germany \\
pavlitska@fzi.de}
}

\maketitle
\thispagestyle{plain}
\pagestyle{plain}

\begin{abstract}
Sparsely-gated Mixture of Expert (MoE) layers have been recently successfully applied for scaling large transformers, especially for language modeling tasks. An intriguing side effect of sparse MoE layers is that they convey inherent interpretability to a model via natural expert specialization. In this work, we apply sparse MoE layers to CNNs for computer vision tasks and analyze the resulting effect on model interpretability. To stabilize MoE training, we present both soft and hard constraint-based approaches. With hard constraints, the weights of certain experts are allowed to become zero, while soft constraints balance the contribution of experts with an additional auxiliary loss. As a result, soft constraints handle expert utilization better and support the expert specialization process, while hard constraints maintain more generalized experts and increase overall model performance. Our findings demonstrate that experts can implicitly focus on individual sub-domains of the input space. For example, experts trained for CIFAR-100 image classification specialize in recognizing different domains such as flowers or animals without previous data clustering. Experiments with RetinaNet and the COCO dataset further indicate that object detection experts can also specialize in detecting objects of distinct sizes.
\end{abstract}


\section{Introduction}

Sparse Mixture of Expert (MoE) layers have recently gained popularity thanks to their ability to scale up models to billions and lately even trillions of parameters~\cite{fedus2022switch,zoph2022designing,du2022glam}. The focus, however, was almost exclusively on transformer models for language modeling tasks. In this work, we insert MoE layers into convolutional neural networks (CNNs) and apply the approach to the basic computer vision tasks of image classification and object detection.  To tackle a well-known problem of unstable expert training, we present soft and hard constraints, encouraging balanced expert utilization. The models further provide an additional hyperparameter to adjust the number of active experts in each forward pass and, thereby, the computational complexity. 

Inherent model interpretability is one side effect of embedding MoE layers into model architectures. For the  language modeling tasks, the experts were shown to mostly specialize on shallow concepts
~\cite{lewis2021base, zoph2022designing, mustafa2022multimodal}. All previous works rely on transformers. To the best of our knowledge, we are the first to study the impact of sparse MoE layers, embedded in CNNs, on model interpretability in computer vision tasks. Due to the larger receptive field, experts can focus on more high-level semantic concepts. 

Our contributions can be summarized as follows:
\begin{itemize}
\item We apply the concept of sparse MoE layers, primarily used in transformers so far, to CNNs.
\item We analyze the semantics of the learned experts and evaluate the impact of different constraints for balancing load specialization on the interpretability strength of the model.
\item We evaluate the concept on two separate tasks: image classification and object detection.
\end{itemize}

\begin{figure}[t]
\centering
\includegraphics[width=\linewidth]{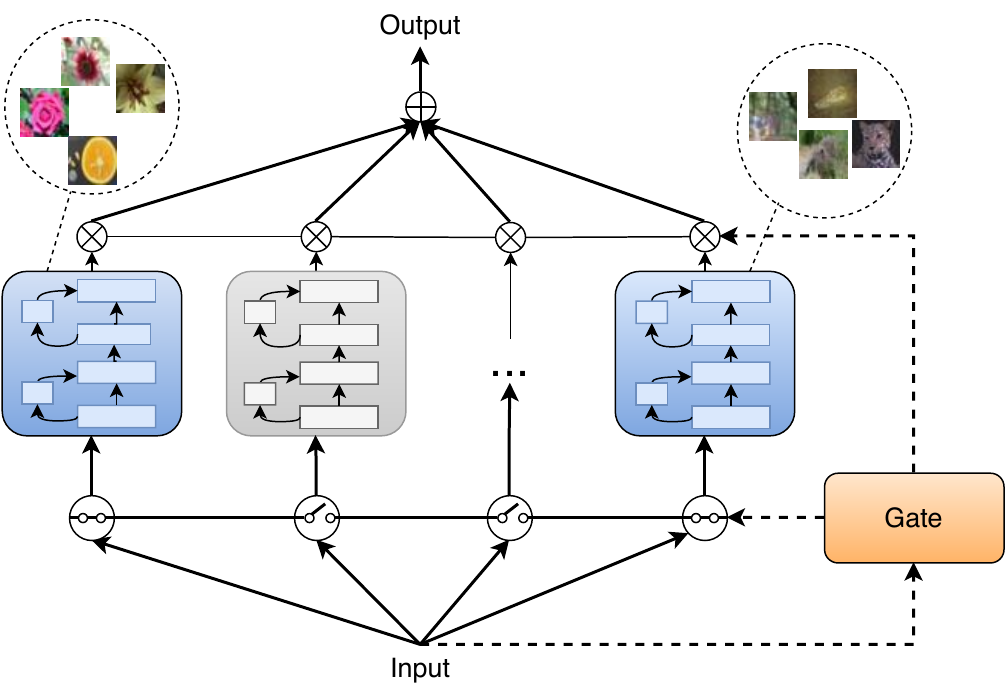}
\caption{The integration of sparse MoE layers into a ResNet-like architecture: each expert is a residual block. The gate selects top $k$ (here: $k=2$) active experts and predicts the weighting of their outputs. When trained, experts specialize on semantic label groups, e.g., flowers or animals.}
\label{fig:moe-blocks}
\end{figure}

\section{Related Work}

\subsection{Sparse MoEs}

Classic mixture-of-expert models, introduced by Jacobs et al.~\cite{jacobs1991adaptive}, consist of a variable number of expert models and a single gate to combine the expert outputs. Eigen et al.~\cite{eigen2014learning} developed the idea of using MoEs as subcomponents of models with individually learned gates. This allows for sharing the remaining parts of architectures and enables multiple MoE layers within a single architecture. The transition to conditional computing through sparse expert activations was first explored by Shazeer et al.~\cite{shazeer2017outrageously} using LSTM networks. By activating a fixed number of experts, it was possible to decouple the number of parameters from FLOPs required for inference through MoEs. 

Fedus et al.~\cite{fedus2022switch} took a further step towards sparse activations and showed that it is possible to train transformer-based MoE models while activating only a single expert, resulting in MoE layers that add only little compute overhead.
Several works~\cite{rajbhandari2022deepspeed,roller2021hash} have since explored the single-expert regime, and Zopf et al.~\cite{zoph2022designing} derived guidelines to design sparse MoEs effectively using transformers.

Sparse MoEs have thus gained popularity for massive language models~\cite{shazeer2017outrageously,lepikhin2021gshard,clark2022unified}. Performance boosts, achieved via sparse MoEs, make training these models possible, although only on large clusters of powerful GPUs. 

In computer vision, the application of sparse MoE has centered around transformer models ~\cite{rajbhandari2022deepspeed,riquelme2021scaling,xue2022go}.  In contrast, in our work, we consider CNNs, which are still the most widely spread architecture in the computer vision area.

Although conventional MoEs at the model level have been successfully evaluated on CNNs \cite{ahmed2016network,gross2017hard,pahuja2019learning,pavlitskaya2020using,pavlitskaya2022evaluating}, embedding MoE layers into CNNs has received significantly less attention. One of the few works is that by Yang et al.~\cite{yang2019condconv}, where the kernels of convolutional layers are activated on a per-example basis. Wang et al.~\cite{wang2019deep} explored a deep dynamic routing approach for CNNs. In their DeepMoE model, each convolutional layer is replaced with an MoE layer, and a multi-headed gating network selects and re-weights the channels in each convolutional layer. DeepMoE is trained end-to-end, it demonstrated a 1000x increase in model capacity while maintaining computational efficiency with only minor losses.

Differently from the work by Wang et al.~\cite{wang2019deep}, we propose experts at the residual block level and analyze to which extent the latent representations learned by a model correspond to human-understandable concepts.

\subsection{Balancing Expert Utilization} 
A known MoE problem is the focus of gates on a small subset of all available experts. 
The weights assigned to other experts are permanently zero or negligibly small. 
Because some experts perform better in the first iterations, the gate increases their probability to be activated. Consequently, these few experts improve above average, and the gate assigns even higher weights to them. This self-reinforcing process continues, s.t. the optimizer ends up in a poor local minimum~\cite{eigen2014learning}.

Eigen et al.~\cite{eigen2014learning} proposed a \textit{hard constraint} on the relative gating assignments to each expert applied during training. 
For this, the weights assigned to each expert are summed over all training samples. 
If this value surpasses the average performance in the first iterations, the gate increases their probability of being zero. 
The remaining positive weights are recomputed using softmax to maintain a convex combination of the experts. Further hard constraints were proposed based on expert  capacities~\cite{fedus2022switch,rajbhandari2022deepspeed,lewis2021base} to encourage the usage of all experts but also to ensure efficient usage of the available hardware.

Shazeer et al.~\cite{shazeer2017outrageously} presented a \textit{soft constraint} approach introducing an auxiliary \textit{importance loss} (see below), which encourages equal importance for all experts during training. The number of training samples per expert may still vary since importance is used instead of the mean number of samples per expert. 

Another auxiliary loss for load balancing was proposed by Lepikhin et al.~\cite{lepikhin2021gshard} -- it limits the number of tokens processed by one expert with a threshold. The auxiliary loss in the switch transformer model~\cite{fedus2022switch} aimed at the uniform distribution of a batch of tokens.

\subsection{Interpretability via Sparse MoEs}

We refer to interpretability as a model ability to explain or to provide meaning to a human in an understandable way~\cite{arrieta2020explainable}. Semantics or visual explainability of a model is crucial to ensure that humans can trust its predictions~\cite{zhang2018interpretable}. Out of a plethora of interpretability approaches~\cite{houben2021inspect}, MoEs favor those which open up the black box and analyze intermediate representations learned by a model.

In the early experiments by Shazeer et al.\cite{shazeer2017outrageously } on the tasks of language modeling, the experts specialized on syntax and/or semantics. The authors provide examples of the specialization for three selected experts: one expert was used for the word \textit{innovation}, another one for the article \textit{a}, and another one on the concept of \textit{fast, rapid} action. 

Lewis et al.~\cite{lewis2021base} showed that experts specialize in very local syntactic information: experts learned clusters of numbers, abbreviations, possessive pronouns, etc. No specialization at the semantic level was observed.

Similar behavior was described in the later work by Zoph et al.~\cite{zoph2022designing}. The experts were found to specialize in punctuation, articles, conjunctions, proper nouns, and numbers. Interestingly, even in the case of multilingual sparse models, no specialization in languages was observed. Instead, the experts continued to focus on the same shallow concepts like punctuation, articles, or numbers. Mustafa et al.~\cite{mustafa2022multimodal} explored multimodal sparsely-activated models. The text experts specialized in nouns and adjectives, whereas image experts specialized on semantic concepts like body parts, textures, fauna, food, and doors. 

The work on visual sparse transformers by Riquelme et al.~\cite{riquelme2021scaling} is closest to ours because it deals with image data. Here, the experts specialize in discriminating between small sets of classes. Expert-class correlation is strong only for the last few layers, whereas no expert specialization was observed for the early layers. Further work with visual transformers by Wu et al.~\cite{wu2022residual} has demonstrated expert specialization across ImageNet classes.

In the case of visual transformers, image patches are tokens, routed to experts. In our work, the routing is performed at the level of whole images, which leads to semantic specialization at the image level.

\section{Approach}
Our approach comprises three components: (1) embedding sparse MoE layers in CNNs, (2) balancing expert utilization via constraints, and (3) expert specialization analysis revealing concepts learned in MoE layers. 

\subsection{Sparsely-gated MoEs for CNNs}

We propose a method to embed sparse MoE layers into CNNs with the goal to achieve computational complexity similar to the baseline in terms of parameters. Without loss of generality, we consider a CNN consisting of residual blocks~\cite{he2016deep} (see Figure \ref{fig:moe-blocks}). The proposed \textit{ResBlock-MoE} architecture uses a complete residual block as its expert. The MoE layer encapsulates multiple replicas of the block that are activated and mixed using a gate. We experiment with inserting MoE layers at different positions in the model architecture. 

\begin{figure}[t]
\centering
\begin{subfigure}[t]{0.4\linewidth}
	\includegraphics[width=0.5\textwidth]{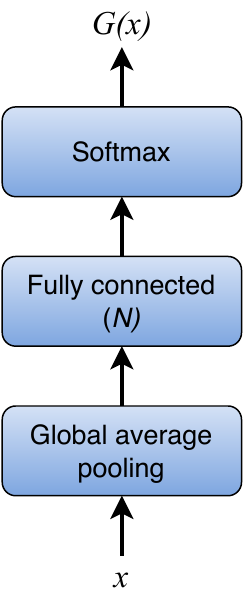}
  	\caption{\textit{GAP-FC} gate}
  	\label{fig:simple-gate}
  \end{subfigure}
 \begin{subfigure}[t]{0.34\linewidth}
  	\includegraphics[width=0.6\textwidth]{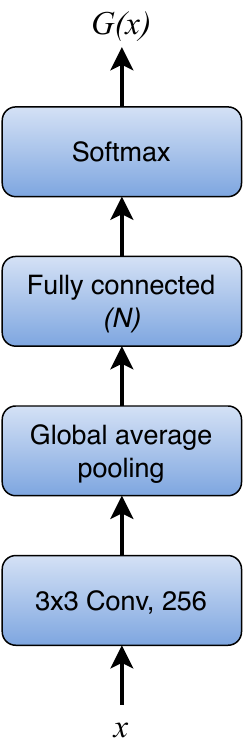}
  	\caption{\textit{Conv-GAP-FC} gate}
  	\label{fig:Conv-GAP-FC}
 \end{subfigure}
\caption{Gate architectures.}
\label{fig:classification_gating_networks}
\end{figure}

We consider two types of gates (see Figure~\ref{fig:classification_gating_networks}): the \textit{GAP-FC} gate consists of a GAP layer followed by a single fully-connected layer, whereas the \textit{Conv-GAP-FC} additionally contains a convolutional layer, which can use detailed, local information encoded in the input features.

\subsection{Constraints to Balance Expert Utilization}

Formally, an MoE consists of a set of $N$ experts $E_1,...,E_N$. For a given input $x$, each expert $E_i$ produces an output $e_i(x)$. 
The gate computes a weight vector $G(x) = [g_1(x),...,g_N(x)]$. The final MoE output is a weighted sum of the expert outputs: $F_{MoE}(x) = \sum_{i=1}^{N}g_i(x)e_i(x)$. To measure the utilization of experts, we define an \textit{importance} vector $I(X)=\sum_{x \in X} G(x)$ for each batch of training samples $X$, and the importance of a single expert $E_i$ as $I_i(X)=\sum_{x \in X} g_i(x)$ \cite{shazeer2017outrageously}. 

We refer to the problem of unbalanced MoE expert utilization as \textit{dying experts}, analogous to the dying ReLU~\cite{lu2019dying}. We consider an expert as dead if it receives less than 1\% average importance on the test set. 
To mitigate the problem, we propose one soft and two hard constraints.

\vspace{1 mm}

\textbf{Hard Constraints}: motivated by the work of Eigen et al.~\cite{eigen2014learning}, we propose two hard constraints on importance. Both hard constraints are only active during training and deactivate experts for an entire batch. 

We denote the mean importance of batch $X$ as $\overline{I(X)}$ and define the \textit{relative importance} of expert $E_i$ for $X$  as follows:
\begin{equation}
I_i^{rel}(X) = \frac{I_i(X) - \overline{I(X)}}{\overline{I(X)}} %= \frac{I_i(X) - \frac{|X|}{N}}{\frac{|X|}{N}}
\end{equation}

In the \textit{relative importance constraint}, the expert weight is zeroed for a batch, if the running relative importance of this expert exceeds the predefined threshold $m_{rel}$: 
\begin{equation}
g_i(X_{t})=0 \iff \sum_{t'=1}^{t-1}I^{rel}_{i}(X_{t'}) > m_{rel}
\end{equation}

For the \textit{mean importance constraint}, we define the mean importance assigned to expert $E_i$ up to time step $t$: 

\begin{equation}
\label{eq:mean_importance}
\overline{I_i(X_t)} = \frac{1}{t}\sum_{t'=1}^{t}\frac{I_i(X_{t'})}{|X_{t'}|}
\end{equation}

In this constraint, the expert weight is zeroed when the mean importance for this expert exceeds the mean importance of the batch by some predefined threshold $m_{mean}$:
\begin{equation}
g_i(X_{t})=0 \iff \overline{I_i(X_{t})} - \overline{I(X_{t})} > m_{mean}
\end{equation}

The relative importance constraint takes a stronger focus on the recent past, whereas the mean importance approach takes a holistic view, with all past importance values having the same impact on the constraint.

\vspace{1 mm}
\textbf{Soft Constraints}: the first soft-constrained approach, originally proposed by Shazeer et al. \cite{shazeer2017outrageously}, is an auxiliary \textit{importance loss} $\mathcal{L}_{imp}$, following Equation \ref{eq:l_importance}. It uses the squared coefficient of variation of importance vector $I(X)$ for batch $X$ and a weighting factor $w_{imp}$:
\begin{equation}\label{eq:l_importance}
\begin{split}
	\mathcal{L}_{imp}=w_{imp}\cdot CV\left( I(X) \right)^2= w_{imp} \cdot \left( \frac{\sigma_{I(X)}}{\mu_{I(X)}} \right) ^2
 \end{split}
\end{equation}

We propose another soft constraint that takes a probabilistic view on expert importance. For this, we interpret an MoE as a probability model in which class probabilities are marginalized over expert selection. 
Each weight $g_i(x)$ is thus the probability $p(E_i|x)$ to select a specific expert for a given input, and outputs of each expert $e_i(x)$ quantify the probability $p(c|E_i,x)$ of each class $c \in C$. 
The MoE output is then defined as follows:
\begin{equation}
\label{eq:fmoe-prob}
F_{MoE}(x) = \sum_{i=1}^{N}p(E_i|x)p(c|E_i,x) = p(c|x)
\end{equation}
%According to Equ.~\ref{eq:fmoe-prob}, 
We interpret the gate output as a discrete probability distribution $P$ with probability $P(E_i|\mathcal{X})$ for expert $E_i$ to be selected for input $\mathcal{X}$ ($\mathcal{X}$ is a random variable for input $x$). 

In expectation, the gate should assign each expert $E_i$ the same average weighting, equal to $\mathbb{E}_X[P(E_i|\mathcal{X})]=\frac{1}{N}$. 
The expected weight assignment thus corresponds to a discrete uniform distribution $Q$ with probability $Q(E_i|\mathcal{X})=Q(E_i)=\frac{1}{N}$.

\begin{table*}[t]
\caption{Expert utilization: coefficients of variation for the number of samples ($CV_{act}$) and importance ($CV_{imp}$) assigned to each expert. A higher $CV$ value means a higher variance in expert utilization. $\#$ stands for the number of  experts alive. We highlight cases when all experts are alive.}
\label{tab:expert-utulization}
\begin{center}
\begin{tabular}{|r | c| c  c  c | c  c  c  | c  c  c  | c  c  c|}
\hline
& MoE & \multicolumn{6}{|c}{Soft constraints} & \multicolumn{6}{|c|}{Hard constraints}\\ 
& position & \multicolumn{3}{|c}{Importance loss} & \multicolumn{3}{|c}{KL-divergence loss} &  \multicolumn{3}{|c}{Relative importance} &  \multicolumn{3}{|c|}{Mean importance}\\ 
&   & $CV_{act}$ & $CV_{imp}$ & $\#$ & $CV_{act}$ & $CV_{imp}$ & $\#$ & $CV_{act}$ & $CV_{imp}$ & $\#$ & $CV_{act}$ & $CV_{imp}$ & $\#$ \\\hline
& 1 & 6.72 & 3.82 & \textbf{4} & 9.23 & 6.36 & \textbf{4} & 56.00 & 56.34 & \textbf{4} & 97.37 & 103.23 & 2.33 \\ 
\textit{ResBlock-MoE}, & 2 & 10.24 & 2.44 & \textbf{4} & 10.55 & 2.99 & \textbf{4} & 52.20 & 51.61 & \textbf{4} & 78.32 & 97.60 & 3.00\\
\textit{4 experts}& 3 & 5.87 & 2.17 & \textbf{4} & 9.35 & 2.21 & \textbf{4} & 43.19 & 25.21 & \textbf{4} & 90.98 & 98.32 & 2.33\\ 
& 4 & 7.97 & 2.86 & \textbf{4} & 8.41 & 2.79 & \textbf{4} & 13.39 & 2.91 & \textbf{4} & 62.99 & 95.53 & 3.00 \\ \hline

% 10 experts
& 1  & 10.69 & 10.43 & \textbf{10} & 14.93 & 13.47 & \textbf{10} & 160.60 & 168.11 & 5.33 & 199.77 & 200.02 & 2.00\\
\textit{ResBlock-MoE}, & 2 & 7.25 & 8.63 & \textbf{10} & 8.62 & 8.56 & \textbf{10} & 165.66 & 186.78 & 5.67 & 170.56 & 182.24 & 3.00\\
\textit{10 experts} & 3  & 6.43 & 5.83 & \textbf{10} & 10.51 & 5.10 & \textbf{10} & 125.13 & 144.25 & 6.67 & 166.65 & 158.01 & 3.33\\
& 4 & 4.80 & 4.48 & \textbf{10} & 8.95 & 6.67 & \textbf{10} & 10.61 & 9.37 & \textbf{10} & 147.22 & 158.82 & 3.67\\ \hline
\end{tabular}
\end{center}
\end{table*}

\begin{table*}[t]
\caption{Mean accuracy and sample standard deviation of image classification models with sparse MoE layers and \textit{GAP-FC} gate.  We highlight cases that beat the baseline accuracy.}
\label{tab:moe-results}
\begin{center}
\begin{tabular}{|r|c|c|c|c|c|}
\hline
& MoE & \multicolumn{2}{|c}{Soft constraints} & \multicolumn{2}{|c|}{Hard constraints}\\ 
 & position  & Importance loss & KL-divergence loss & Relative importance & Mean importance \\\hline
 
\textit{Baseline} & & \multicolumn{4}{|c|}{72.62$\pm$0.29} \\\hline

 %% 4 experts
& 1& 72.24$\pm$0.49 & \textbf{72.72}$\pm$0.36   & 72.21$\pm$0.42 & \textbf{73.00}$\pm$0.40\\
\textit{ResBlock-MoE}, & 2 &  72.18$\pm$0.29 & 72.25$\pm$0.17  & 72.18$\pm$0.48 & \textbf{72.95}$\pm$0.35\\
\textit{4 experts} & 3 & 71.65$\pm$0.43 & 71.54$\pm$0.22 & 72.05$\pm$0.15 & 72.61$\pm$0.37 \\
& 4 &  71.95$\pm$0.39 & 71.80$\pm$0.23 & \textbf{73.10}$\pm$0.25 & 72.57$\pm$0.31  \\\hline

 %% 10 experts
& 1& 71.51$\pm$0.23 &  71.32$\pm$0.60 & 71.60$\pm$0.09 & \textbf{72.28}$\pm$0.15 \\
\textit{ResBlock-MoE,} & 2 &  \textbf{72.76}$\pm$0.50 & 71.87$\pm$0.30  & 71.51$\pm$0.44 &\textbf{72.08}$\pm$0.32\\
\textit{10 experts}& 3 & 71.47$\pm$0.29  & 71.43$\pm$0.12  & 70.82$\pm$0.16 & \textbf{72.05}$\pm$0.61 \\ 
& 4 &  71.61$\pm$0.16 &  71.99$\pm$0.23 & 72.84$\pm$0.30 & \textbf{73.09}$\pm$0.35 \\ \hline
\end{tabular}
\end{center}
\end{table*}

We define an auxiliary \textit{KL-divergence loss} $\mathcal{L}_{KL}$ as the KL-divergence $D_{KL}(P||Q)$ between $P$ and $Q$, weighted by hyperparameter $w_{KL}$. 
The probability $P(E_i|\mathcal{X}=X)=\frac{I_i(X)}{|X|}$ is computed as the average importance per sample in batch $X$. 
$\mathcal{L}_{KL}$ is then defined as follows:
\begin{equation}
\label{eq:KL-loss}
\begin{split}
    \mathcal{L}_{KL} = w_{KL} \cdot D_{KL}(P||Q) =\\
    w_{KL} \cdot \sum_{i=1}^{N}P(E_i|X) \cdot \ln\Big(\frac{P(E_i|X)}{Q(E_i)}\Big)\\
    = w_{KL} \cdot \sum_{i=1}^{N}\frac{I_i(X)}{|X|} \cdot \ln\Big(\frac{I_i(X) \cdot N}{|X|}\Big)
\end{split}
\end{equation}

$\mathcal{L}_{imp}$ penalizes inequality in importance distribution harder than $\mathcal{L}_{KL}$, thus achieving an equal expert utilization. On the other hand, $\mathcal{L}_{KL}$ leads to higher variance in the expert utilization but still avoids dying experts.

\section{Experiments with Sparse MoEs for Image Classification}

\subsection{Experimental Setup} 
We use the ResNet-18~\cite{he2016deep} architecture\footnote{Available at \url{https://github.com/pytorch/vision/blob/master/torchvision/models/resnet.py}} consisting of four ResNet blocks, while the first convolutional and pooling layers were replaced by a single 3x3 convolution to adjust for the lower input resolution. We run experiments on the CIFAR-100 dataset~\cite{krizhevsky2009learning}. The MoE experts follow the architecture of the residual blocks, but we adjust the number of filters of each expert in a bottleneck manner, thus reducing the number of parameters to maintain comparable computational complexity. 
We also add an additional projection shortcut that connects the MoE layer input with its output. We evaluate embedding sparsely-gated MoE in each possible position in ResNet-18. Since representations learned by a CNN evolve from earlier to later layers, we expect the MoE layer to learn human-understandable concepts to a greater extent when inserted in later layers.

We train models with 4 experts and set the number of active experts to $k=2$. We set weights for auxiliary losses to $w_{imp}=w_{KL}=0.5$, and thresholds for hard constraints to $m_{rel}=0.5$ and $m_{mean}=0.3$. 

\begin{figure*}[t]
 \centering
 \begin{subfigure}[b]{0.24\linewidth}
  	\includegraphics[width=\textwidth]{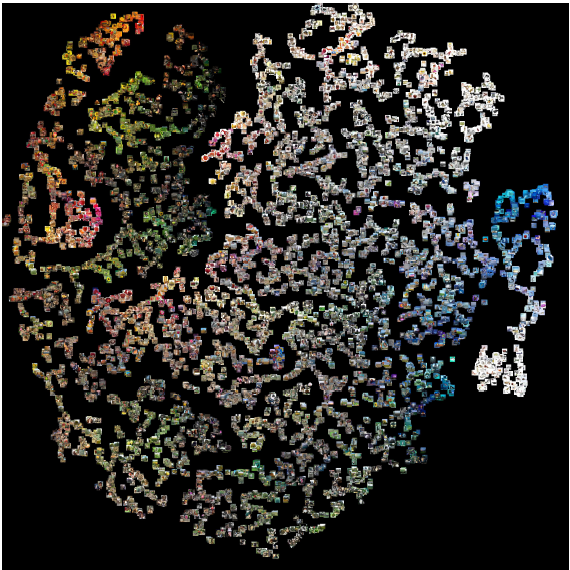}
  	\caption{Importance loss,  pos. 1}
  	\label{fig:tsne-moe-imp-1-4}
 \end{subfigure}
  \begin{subfigure}[b]{0.24\linewidth}
  	\includegraphics[width=\textwidth]{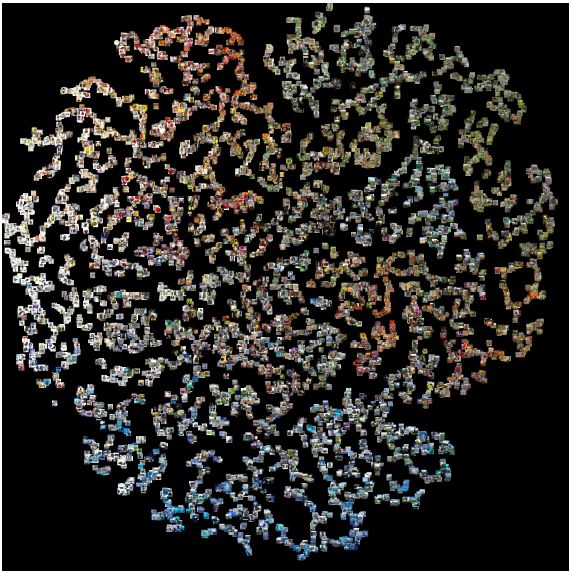}
  	\caption{Importance loss,  pos. 4}
  	\label{fig:pca-moe-imp-1-4}
 \end{subfigure}
 \begin{subfigure}[b]{0.24\linewidth}
  	\includegraphics[width=\textwidth]{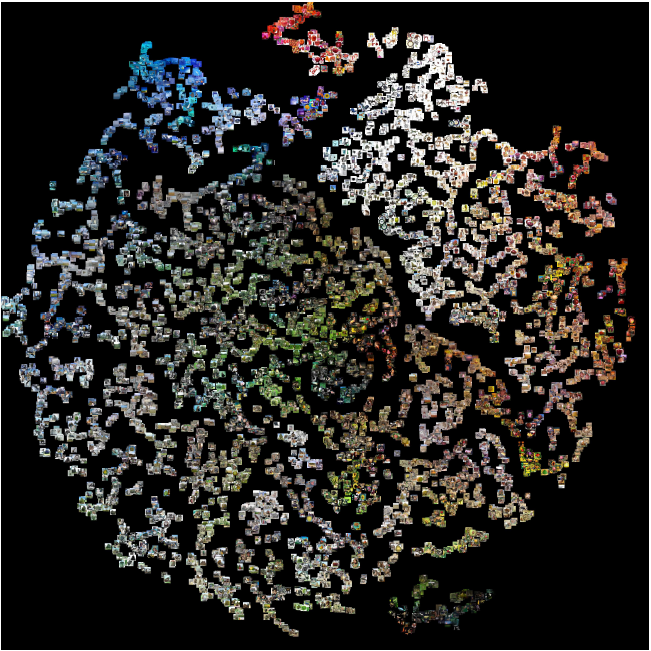}
  	\caption{Relative importance,  pos. 1}
  	\label{fig:tsne-moe-imp-4-4}
 \end{subfigure}
  \begin{subfigure}[b]{0.24\linewidth}
  	\includegraphics[width=\textwidth]{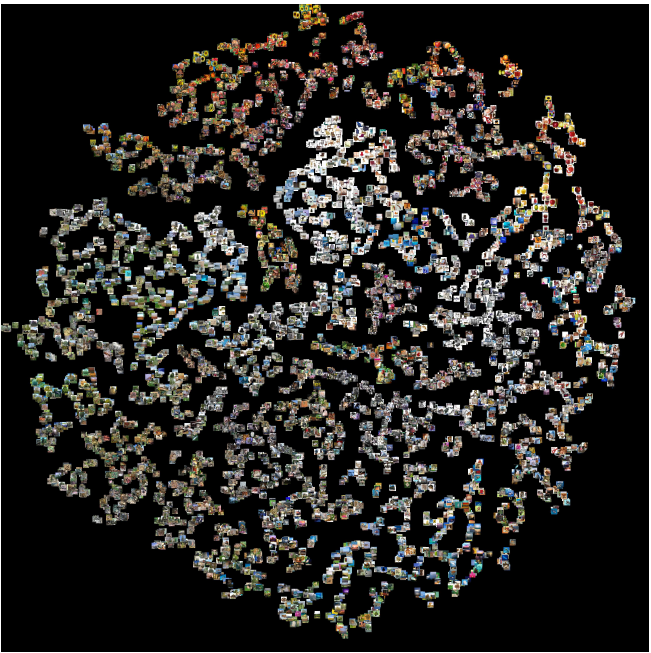}
  	\caption{Relative importance,  pos. 4}
  	\label{fig:pca-moe-imp-4-4}
 \end{subfigure}

 \caption{Visualization of gate logits using t-SNE for \textit{ResBlock-MoE} with 4 experts and a \textit{GAP-FC} gate.}
 \label{fig:viz-gatelogits}
 \end{figure*}

\begin{figure*}[t]
\begin{center}
 
 \begin{subfigure}[b]{0.32\linewidth}
  	\includegraphics[width=\textwidth]{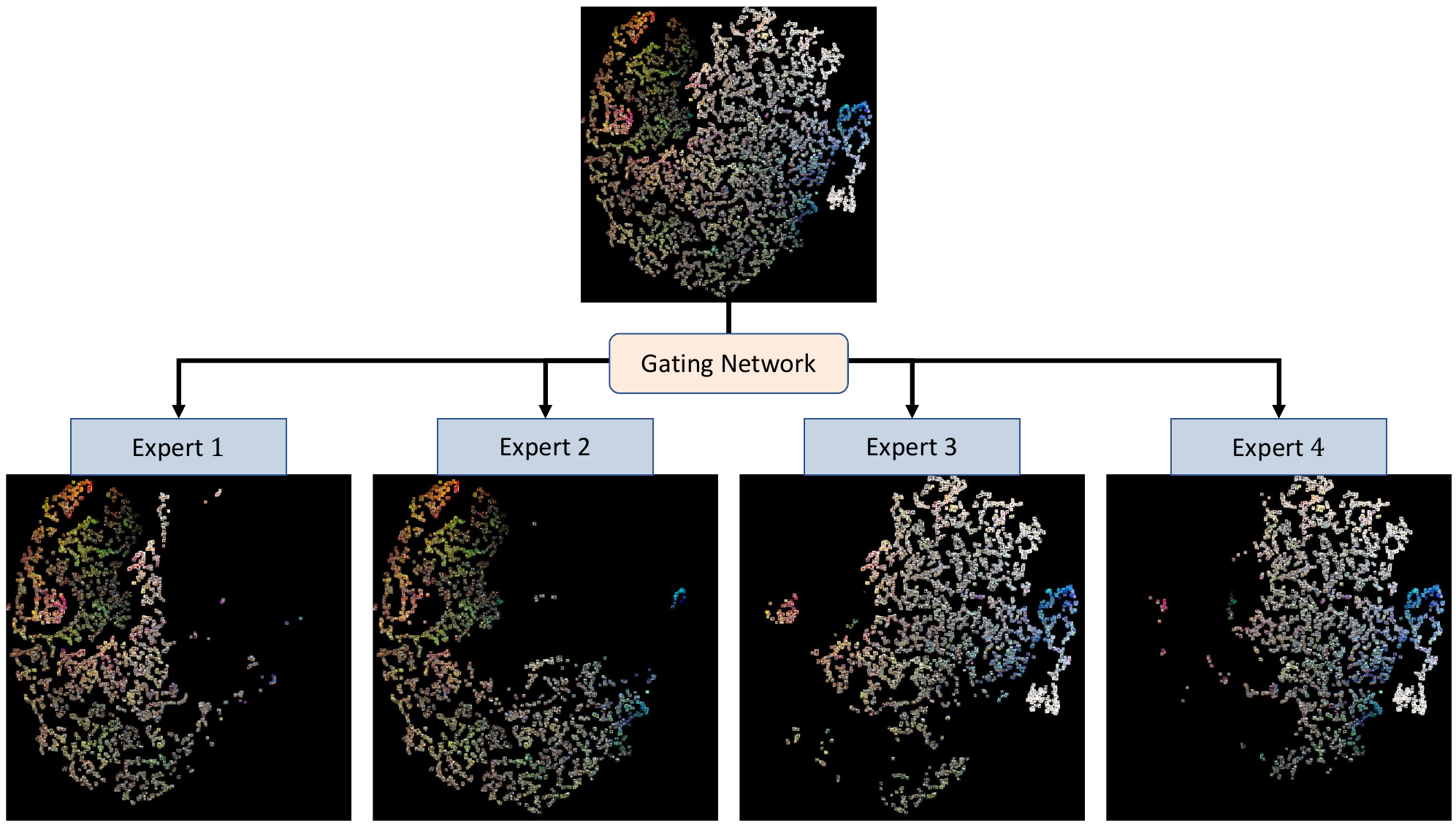}
  	\caption{\textit{GAP-FC} gate, importance loss, MoE layer at position 1}
 \end{subfigure}
 \begin{subfigure}[b]{0.32\linewidth}
  	\includegraphics[width=\textwidth]{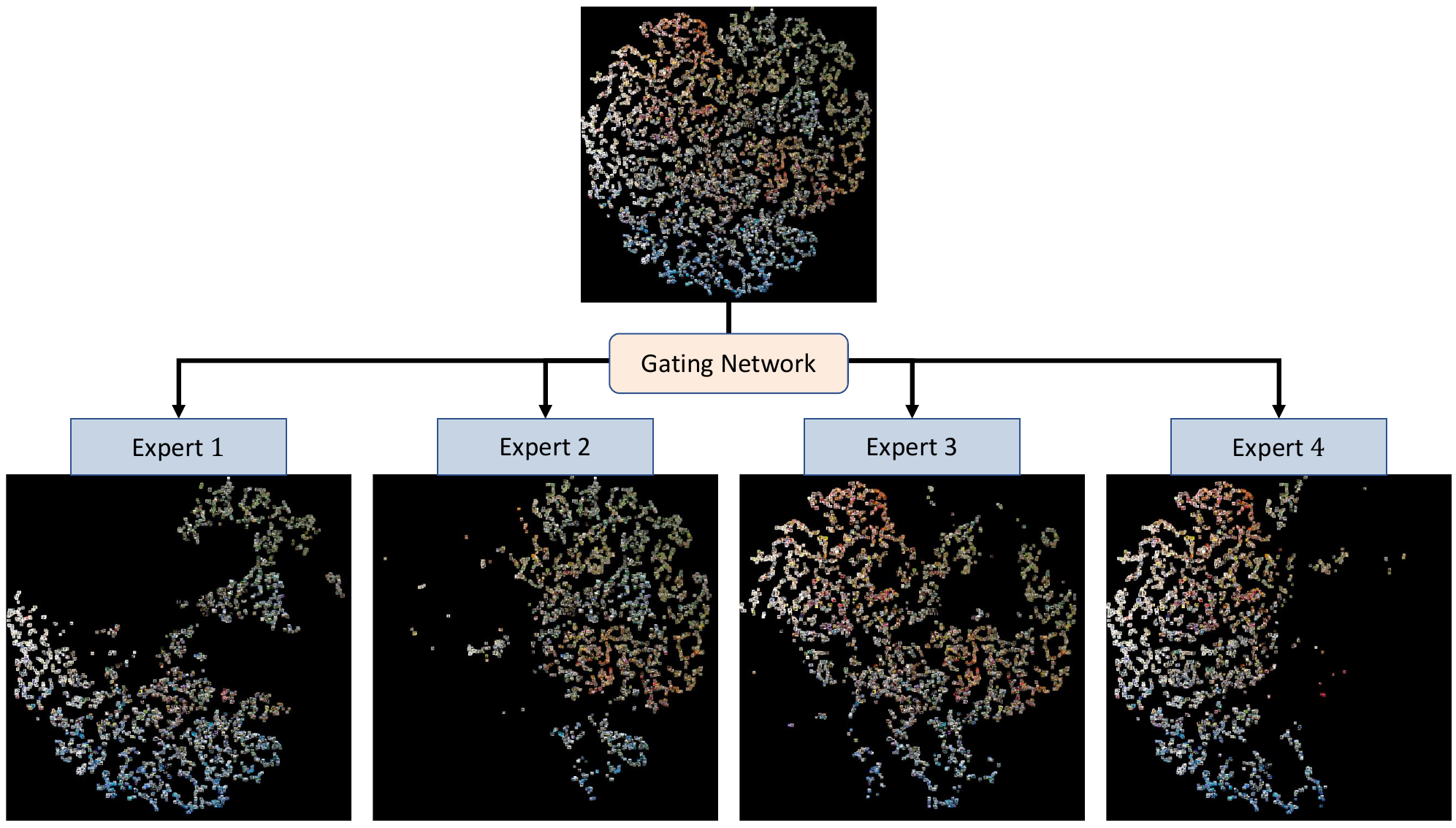}
  	\caption{\textit{GAP-FC} gate,  importance loss, MoE layer at position 4}
 \end{subfigure}
     \begin{subfigure}[b]{0.32\linewidth}
  	\includegraphics[width=\textwidth]{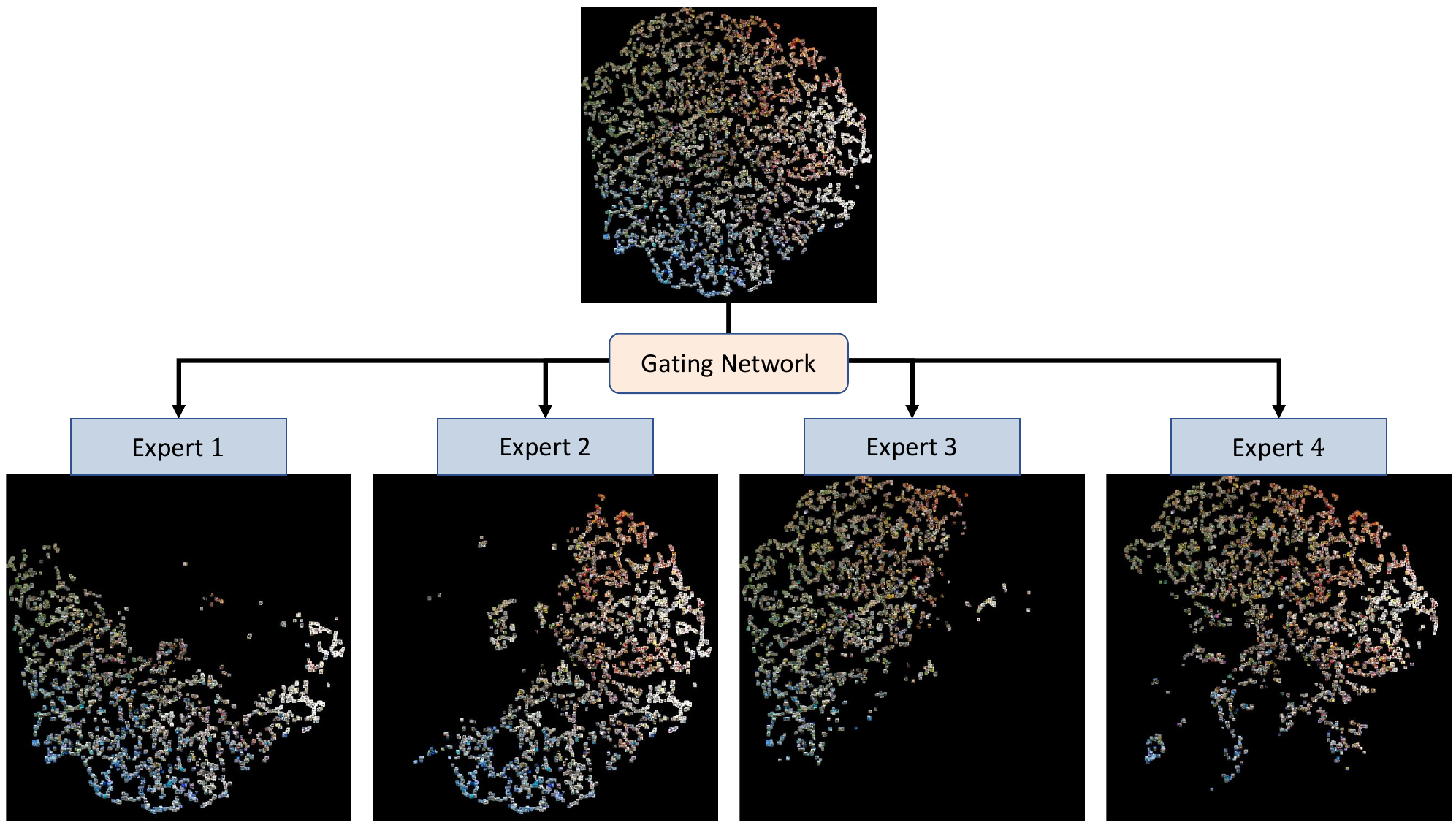}
  	\caption{\textit{Conv-GAP-FC} gate, importance loss, MoE layer at position 4}
  	\label{fig:tsne-gate-moe-imp-1-4}
 \end{subfigure}
 
 \begin{subfigure}[b]{0.32\linewidth}
  	\includegraphics[width=\textwidth]{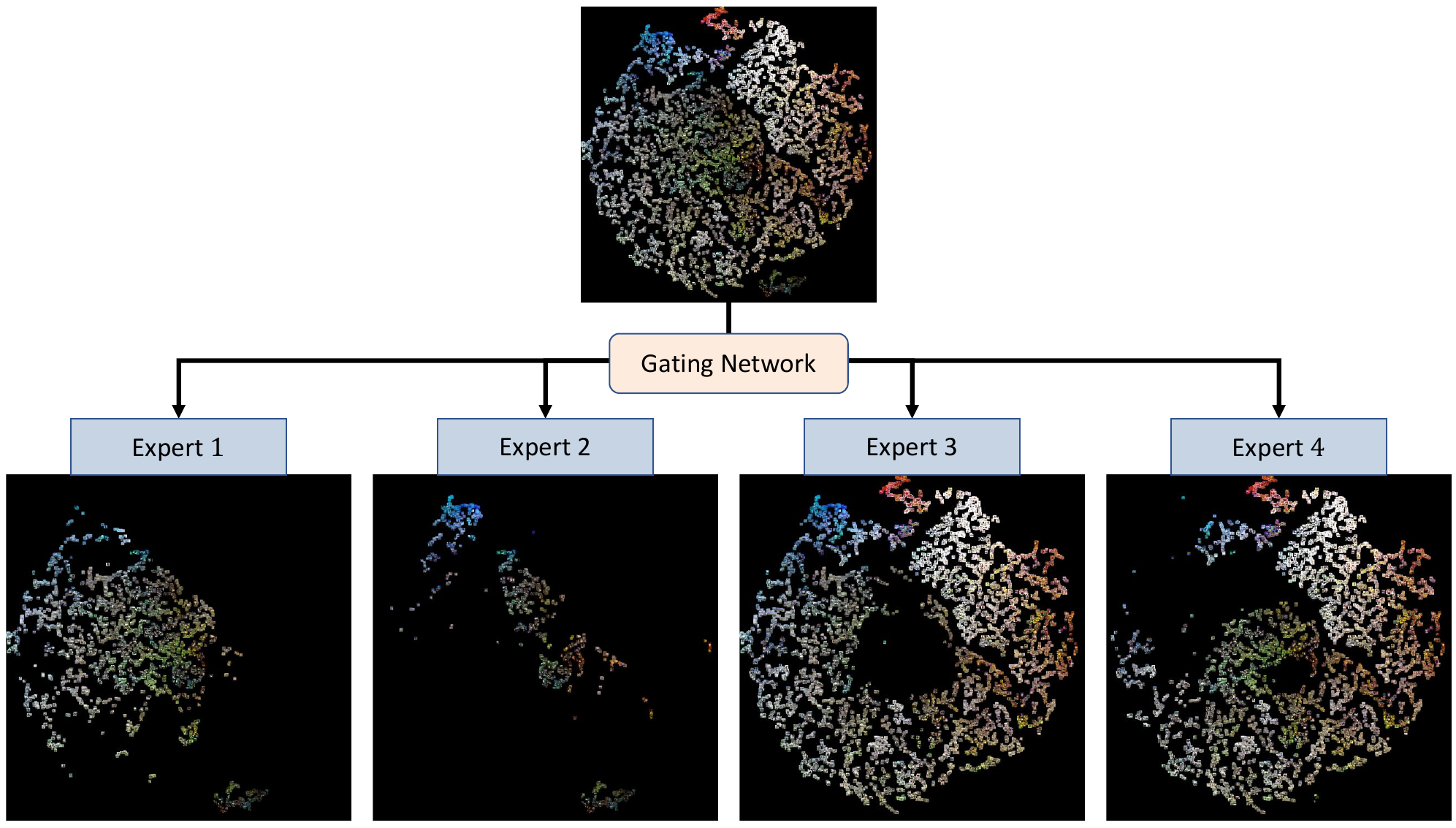}
  	\caption{\textit{GAP-FC} gate, relative importance constraint, MoE layer at position. 1}
 \end{subfigure}
  \begin{subfigure}[b]{0.32\linewidth}
  	\includegraphics[width=\textwidth]{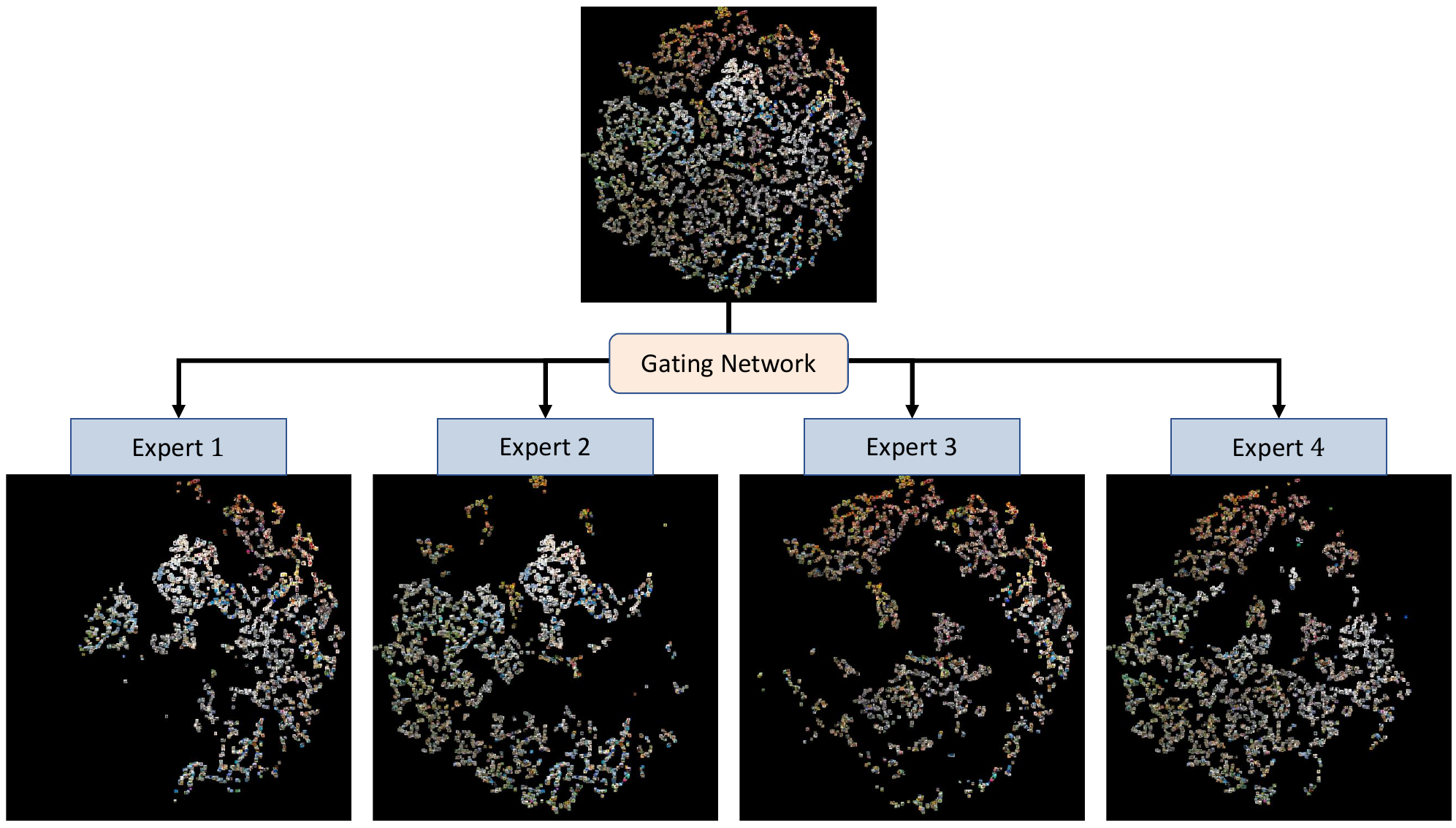}
  	\caption{\textit{GAP-FC} gate, relative importance constraint, MoE layer at position 4}
 \end{subfigure}
 \caption{Assignment of input samples to specific experts in models with \textit{ResBlock-MoE} with 4 experts, visualized with t-SNE.}
 \label{fig:viz-tsne-distr}
 \end{center}
 \end{figure*} 

All experiments are performed using NVIDIA GeForce RTX 2080 TI GPU.  The models are trained with batch size 128 for 150 epochs with the Adam optimizer~\cite{kingma2014adam}.  All experiments are repeated three times, and the averaged values are reported. 

\subsection{Performance and Expert Utilization }
\textbf{Dying experts}: soft constraints successfully mitigated the experts death (see Table \ref{tab:expert-utulization}).  Both auxiliary losses minimize the variation in the importance per expert, whereas a higher variation in the number of samples per expert was observed for the KL-divergence loss. Hard constraints, however, demonstrated worse results. Mean importance models could not keep all experts alive in a single model, whereas relative importance models had no dying experts for models with 4 experts. Also, increasing the number of experts to 10 led to more variance in expert utilization. 

\textbf{Accuracy and computational cost}: in our setting, embedding MoE layers into the model aims at boosting the interpretability strength, not at beating the baseline performance. Out of the evaluated models, the best results were achieved with the mean importance constraint (see Table \ref{tab:moe-results}). The choice of a constraint thus provides an evident trade-off between overall performance and the death of experts.

In all models, we keep the number of active experts $k=2$ to maintain a computational budget comparable to the baseline. We measure the computational budget in the multiply–accumulate (MAC) operations, and the baseline reaches 0.56 GMac. Although increasing the number of active experts $k$ led to better performance, each additionally activated expert adds about 0.06 to 0.08 GMac depending on the layer position. For $k=3$, the overall model reaches on average 0.63 GMac, for $k=4$ already 0.7 GMac. $k$ thus controls a trade-off between accuracy and computational complexity. 

\textbf{Impact of the gate architecture}: we have evaluated replacing the \textit{GAP-FC} gate with \textit{Conv-GAP-FC} gate, exemplary for \textit{ResBlock-MoE} with 4 experts at position 4 and importance loss. This model reached the accuracy of 72.42 $\pm$ 0.27, beating the corresponding \textit{GAP-FC} model (71.95$\pm$0.39), but not the baseline (72.62$\pm$0.29). Hard-constrained models using \textit{Conv-GAP-FC} gate suffer massively from dying experts, even with a decreased learning rate. 

\begin{table*}[t]
\caption{Class-wise weight allocation for the ResNet-18 models. Results for the top 5 classes for which the experts receive the largest weights.}
\label{tab:weight-allocation}
\begin{center}
\begin{tabular}{|r|r | r | c| l  c |l c|l c|l c|}
\hline
\multicolumn{2}{|c}{Constraint} & Gate & MoE & \multicolumn{2}{c}{Expert 1} & \multicolumn{2}{|c}{Expert 2} & \multicolumn{2}{|c}{Expert 3} & \multicolumn{2}{|c|}{Expert 4} \\ 
\multicolumn{2}{|c}{ } && position & Class & Weight & Class & Weight & Class & Weight & Class & Weight  \\ \hline

\multirow{15}{*}{\rotatebox[origin=c]{90}{\parbox[c]{5cm}{\centering Soft }}} & \multirow{15}{*}{\rotatebox[origin=c]{90}{\parbox[c]{5cm}{\centering Importance loss}}} & \multirow{10}{*}{\rotatebox[origin=c]{90}{\parbox[c]{2cm}{\centering \textit{GAP-FC}}}} &  & orange & 0.71  & caterpillar & 0.4 & chair & 0.59 &  dolphin & 0.78  \\
&& && rose & 0.64 & forest & 0.38 &  cockroach & 0.55  & shark & 0.68  \\
& && 1& apple & 0.59 & mushroom & 0.34  & tank & 0.46  & mountain & 0.66  \\
& &&& sweet pepper & 0.58  & butterfly & 0.31 & plate & 0.45  & skyscraper & 0.65  \\
& &&& sunflower & 0.56  & aquarium fish & 0.30 & lawn mower & 0.43 & sea & 0.63 \\\cline{4-12}

& && & mountain & 0.86 & leopard & 0.78& poppy & 0.80 & chair & 0.85 \\
&&& & dolphin & 0.86 & forest & 0.75 & orange & 0.79 & cockroach & 0.85  \\
&&& 4 & shark & 0.80 & porcupine & 0.72 & rose & 0.76 & telephone & 0.65  \\
&&&& sea & 0.79  & tiger & 0.64  & tulip & 0.71  & lamp & 0.64  \\
&&&& whale & 0.79  & mushroom & 0.63  & sunflower & 0.71  & bottle & 0.63  \\\cline{3-12}

&& \multirow{5}{*}{\rotatebox[origin=c]{90}{\parbox[c]{1.5cm}{\centering \textit{Conv-GAP-FC}}}} &  & plain & 0.88 & chair & 0.90&  leopard & 0.77 &  poppy & 0.74  \\
&&&& mountain & 0.87  & telephone & 0.80  & porcupine & 0.75  & orange & 0.70  \\
&&&4 & sea & 0.84  & cockroach & 0.78  & crocodile & 0.65  & tulip & 0.69  \\
&&&& dolphin & 0.82  & clock & 0.63  & forest & 0.64  & rose & 0.66  \\
&&&& cloud & 0.78  & cup & 0.60  & kangaroo & 0.63  & hamster & 0.65\\\hline

\multirow{5}{*}{\rotatebox[origin=c]{90}{\parbox[c]{1.5cm}{\centering Hard}}} &\multirow{5}{*}{\rotatebox[origin=c]{90}{\parbox[c]{1.5cm}{\centering Relative importance}}} & \multirow{5}{*}{\rotatebox[origin=c]{90}{\parbox[c]{2cm}{\centering \textit{GAP-FC}}}} & & chair & 0.96  & plain & 0.87& poppy & 0.86  & wolf & 0.73 \\
&&&& television & 0.79  & willow tree & 0.82  & sunflower & 0.83  & racoon & 0.69  \\
&& & 4 &bottle & 0.79  & oak tree & 0.79 &   rose & 0.80 &  possum & 0.60  \\
&&&& telephone & 0.75  & forest & 0.68  & tulip & 0.77 & motorcycle & 0.58 \\
&&&& cup & 0.74  & maple tree & 0.66  & orange & 0.74  & skunk & 0.56  \\\hline

\end{tabular}
\end{center}
\end{table*}

\begin{table*}[h]
    \caption{Correlation between weight assignment and expert performance for the models with the \textit{ResBlock-MoE} layer at position 4.}
    \label{tab:correlation}
    \centering
    \begin{tabular}{|l|c c | c c|}
    \hline
           & \multicolumn{2}{c|}{Importance loss} & \multicolumn{2}{c|}{Relative importance constraint} \\
          & Pearson  & Spearman & Pearson & Spearman \\\hline
         Correlation between expert accuracy and sparse weights  &  0.8720 &  0.8870 & 0.5827 &  0.5765 \\
         Correlation between expert accuracy and non-sparse weights &  0.8705 & 0.8903  & 0.5919 &  0.5981\\
         Correlation between expert accuracy and activations per expert &  0.8501 & 0.8301  & 0.5981 & 0.6172\\\hline
    \end{tabular}

\end{table*}
 
\subsection{Interpretability via Sparsely-Gated MoEs}

\textbf{Dataset partitioning by gate}: visual assessment of the gate logits, plotted with t-SNE (see Figure~\ref{fig:viz-gatelogits}) demonstrates,
that MoE at position 1 leads to assignment based on the dominant colors of input images, whereas for position 4, the distinctions are much more faded. Furthermore, visible structures are less significant for the soft-constrained models, compared to the hard-constrained case. 

The resulting sample assignment to different experts reveals more striking differences across constraints (see Figure~\ref{fig:viz-tsne-distr}). For earlier layers, the gate divides the data into 2 major subdomains, while for position 4, the gate varies more between different expert combinations. Weight assignment in deeper MoE layers is thus based more on high-level features and leads to stronger differentiation. For hard-constrained models, visible structures are less significant.

A more complex \textit{Conv-GAP-FC} gate subdivides the assignments clearer. The gate thus produces less unambiguous weight vectors and selects experts definitely.

\textbf{Experts specialization}: to analyze the implicit specialization of experts on distinct subdomains of the input space, we only activate one specific expert during evaluation and assign all weights to it. We then analyzed classes that received the largest weights in each expert during evaluation (Table \ref{tab:weight-allocation}). We could observe distinct repeating clusters of classes for different models, e.g., flowers, marine animals, trees, and furniture. MoE layers thus learn the natural clustering of the concepts represented in input images. Repeating clusters are formed regardless of the gate architecture and constraint, but inserting MoE layers at the deeper positions in a network led to larger weights assigned to the experts, indicating better specialization.

Furthermore, we determined, that a gate chooses the best-performing expert for the images of the corresponding cluster.  For this, we extracted results for classes in which each expert is assigned the highest and lowest weights. Full MoE performs in 73 out of 100 classes at least as well as its best experts in this domain. The gate can thus identify experts for distinct domains and suitably support experts with inferior performance. It is consequently able to reasonably combine the output feature maps to improve the overall predictions. 

\textbf{Expert utilization vs. accuracy}: next, we evaluated the correlation between the average weights assigned to an expert (sparse and non-sparse, i.e. with all experts activated) and every single expert's test accuracy per class. We also evaluated the correlation between accuracy and the number of activations per expert. The results (see Table \ref{tab:correlation}) indicate a strong relationship for the soft-constrained model. For the hard-constrained case, the experts are more generalized and do not show as large performance variations on different classes, the gate does not rely on the same experts for a certain class.

\section{Experiments with Sparse MoEs for Object Detection}

\subsection{Experimental Setup} 
We use a pre-trained\footnote{Available at \url{https://github.com/yhenon/pytorch-retinanet}} \textit{RetinaNet}~\cite{LinGGHD17} with the ResNet-50 backbone as a baseline, and the COCO dataset~\cite{lin2014microsoft}. We train all models using $\gamma=2$ and $\alpha=0.25$ for the focal loss. %We report standard COCO metrics \mbox{mAP@[.5, .95]} and mAP@.5.

We embed the sparse MoE layers in two manners: (1) \textit{2Block-MoE}: by replacing the regression and classification subnets with two separate MoE blocks, and (2) \textit{SingleMoE}: with a single gate shared between regressor and classifier. We keep the backbone weights frozen during training. We also trained models with unfrozen weights but did not observe performance improvements. Mean importance constraints are not included in the evaluation, since the models suffer from massive dying expert problems.  

The gate is \textit{Conv-GAP-FC} (see Figure \ref{fig:Conv-GAP-FC}). All experts are initialized using the Kaiming approach~\cite{he2015delving} to learn more diverse features. Weighting factors are reduced to $w_{imp}=w_{KL}=0.25$ to guarantee better expert utilization for deeper MoE layers. Additionally, we set $m_{rel}=0.3$ to avoid dying experts. We train models with 4 experts and set $k=2$.

\subsection{Evaluation}
\textbf{Performance}: the hard-constrained models performed slightly better than the soft-constrained ones, although none of the models with MoE layers outperformed the baseline (see Table \ref{tab:DetectorMoE}). We further observe only a slight performance drop for the \textit{Single} model compared to two separate gates.

\begin{table}[t]
\caption{Object detection mAP (\%) of MoE models on COCO test-dev2017. We
highlight cases that beat the baseline accuracy.}
\label{tab:DetectorMoE}
\begin{center}
\begin{tabular}{|r | r|cc|}
\hline
Model & Constraint & mAP & mAP@.5  \\
& &@[.5, .95] & @.5 \\\hline
\textit{Baseline} & & 35.0 & 52.5  \\ \hline
\textit{SingleMoE} & KL-divergence loss & 33.4 & 50.9 \\ \hline
& KL-divergence loss & 33.5 & 50.9  \\
\textit{2Block-MoE}&  Importance loss & 33.5 & 50.9  \\ 
& Rel. imp. constraint & 33.7 & 51.0  \\ \hline
\textit{2Block-MoE}  & KL-divergence loss  & 33.6 & 50.9 \\ 
with Conv4 & Rel. imp. constraint & 33.6 & 50.9 \\ \hline 
\textit{2Block-MoE} with & KL-divergence loss & \textbf{35.1} & \textbf{52.7} \\ 
pre-trained weights & Rel imp. constraint & 34.4 & 51.7 \\\hline
\textit{2Block-MoE}  with Conv4 &  & & \\ 
 and pre-trained weights & KL-divergence loss & \textbf{35.1} & \textit{52.6}\\ \hline
\end{tabular}

\end{center}
\end{table}

\textbf{Specialization of the \textit{2Block-MoE} in regressor}:  to gain insights into the weighting decisions of BBox regressor, we first analyze the behavior visually. For this, we keep $k_{cls}=2$ experts active in the classifier MoE and only analyze the regressor MoE. 

\begin{figure}[t!]
    \centering
      \resizebox{1.0\linewidth}{!}{
    \begin{tabular}{c c}
     \includegraphics[width=0.6\linewidth]{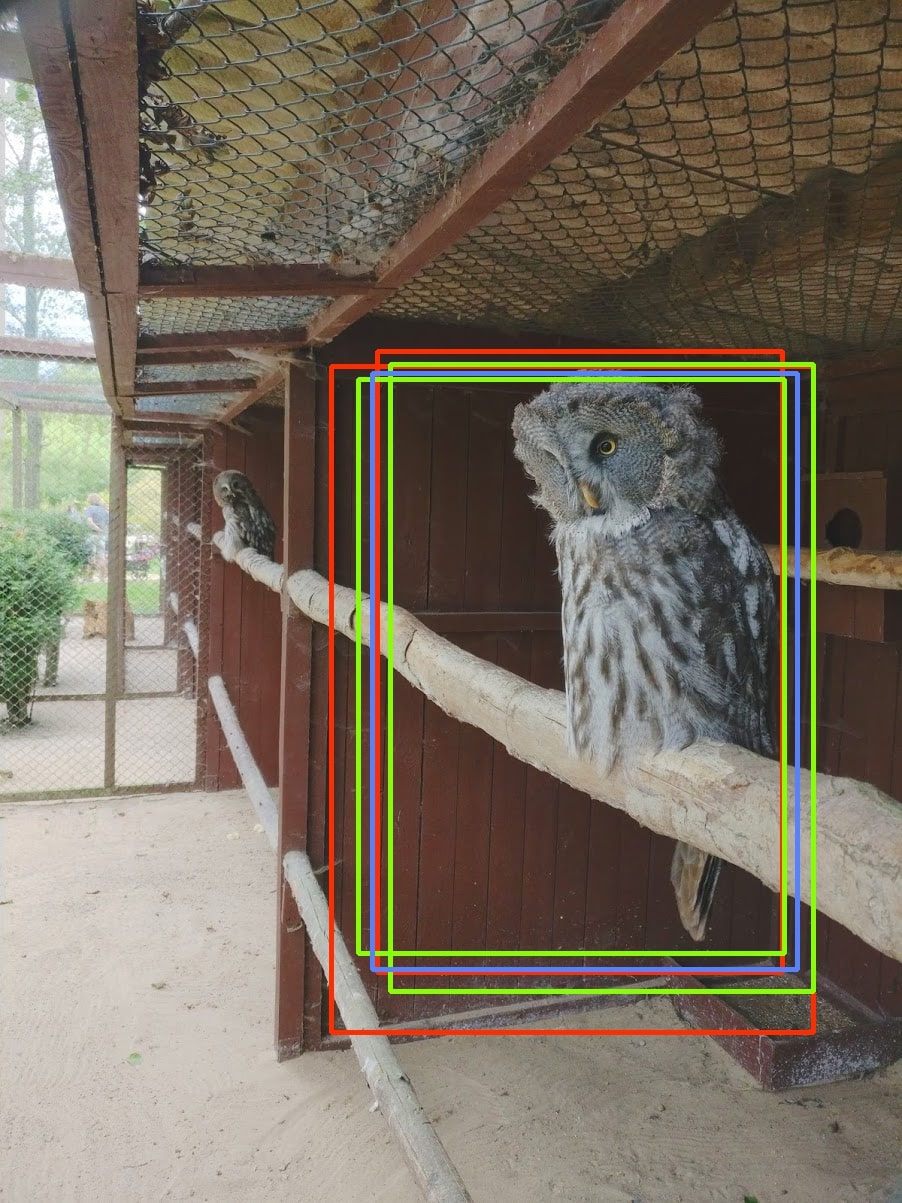}    &  \includegraphics[width=0.6\linewidth]{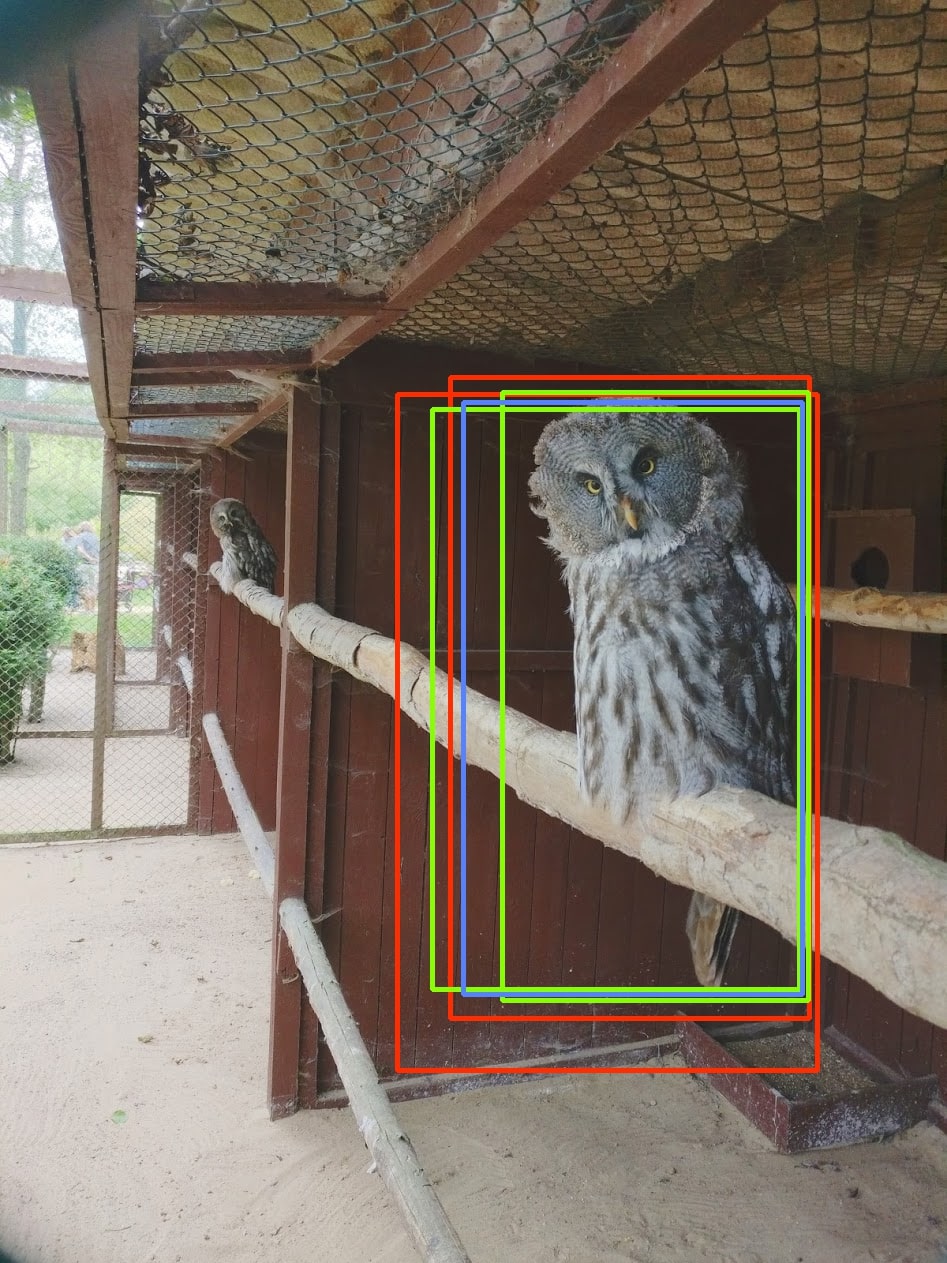}\\
      (a) KL-divergence loss & (b) KL-divergence loss \\
      Weights: (\textbf{0.44}, 0.15, 0.19, \textbf{0.22})   & Weights: (\textbf{0.44}, 0.15, 0.19, \textbf{0.22}) \\ 
      Sparse weights: (0.67, 0.33) &  Sparse weights: (0.67, 0.33) \\ 
        \\
     \includegraphics[width=0.6\linewidth]{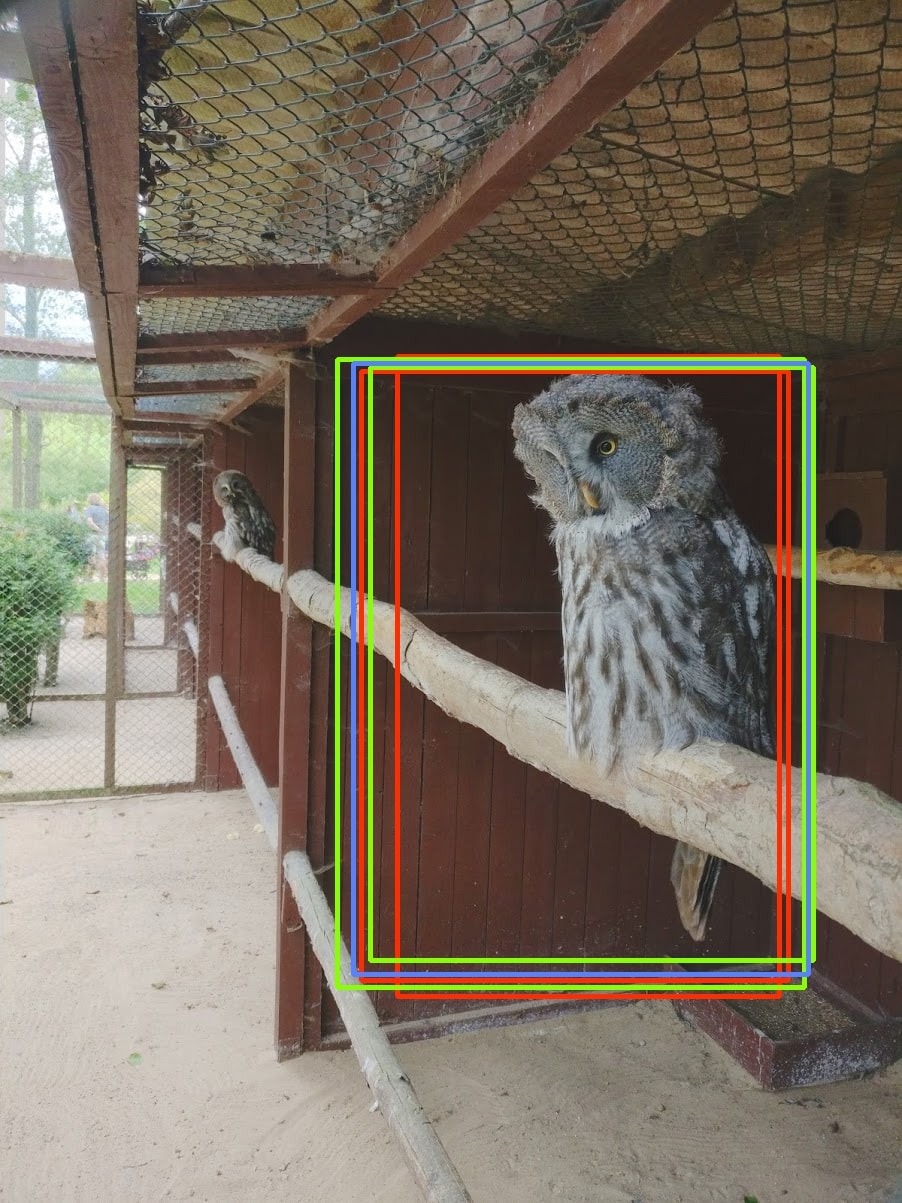}    &  \includegraphics[width=0.6\linewidth]{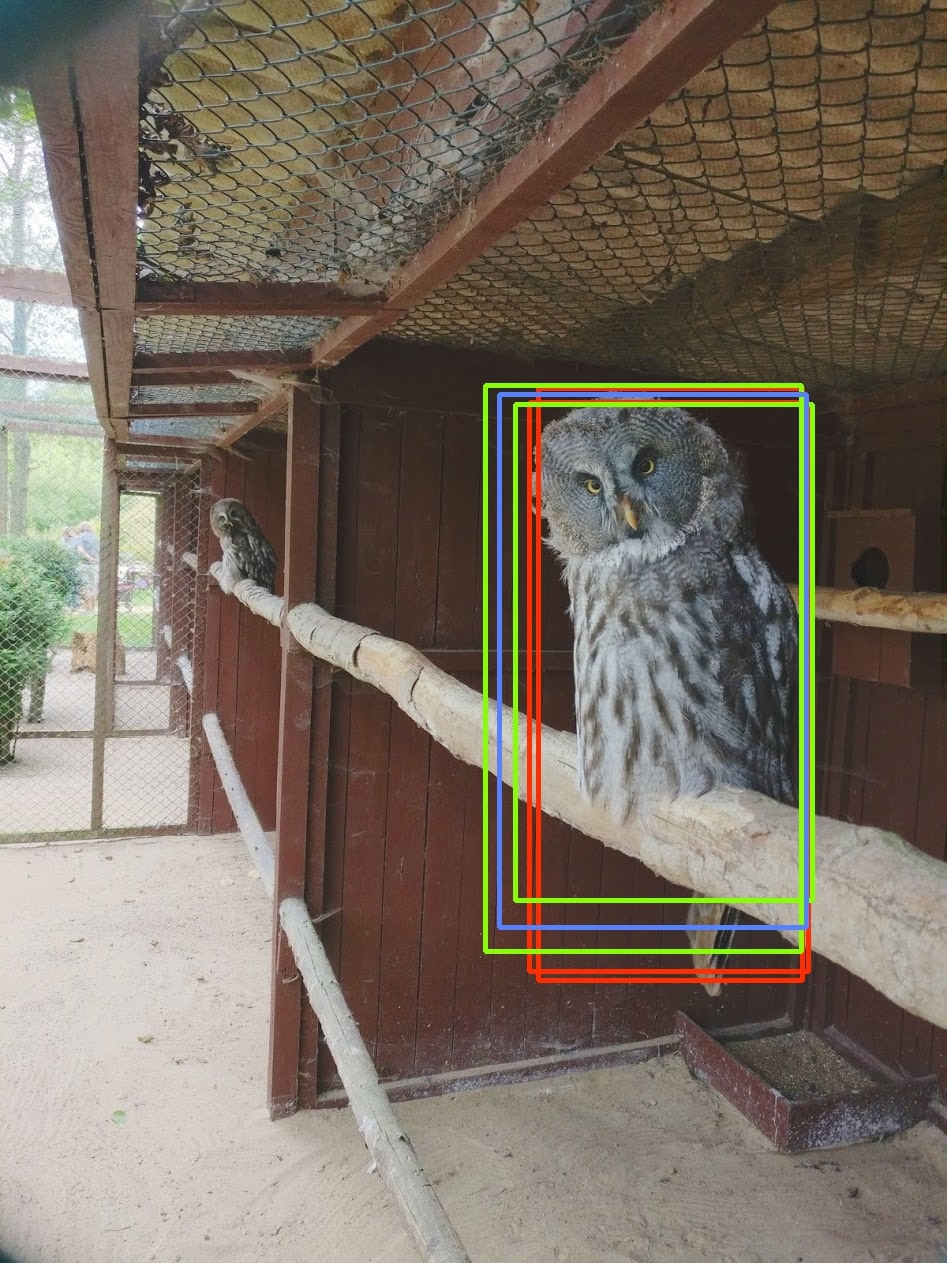}\\
      (c) Importance loss & (d) Importancee loss \\
      Weights: (\textbf{0.29}, \textbf{0.33}, 0.21, 0.16)   & Weights: (0.18, 0.15, \textbf{0.23}, \textbf{0.44}) \\ 
      Sparse weights: (0.47, 0.53) &  Sparse weights: (0.35, 0.65) \\
        \\
       \includegraphics[width=0.6\linewidth]{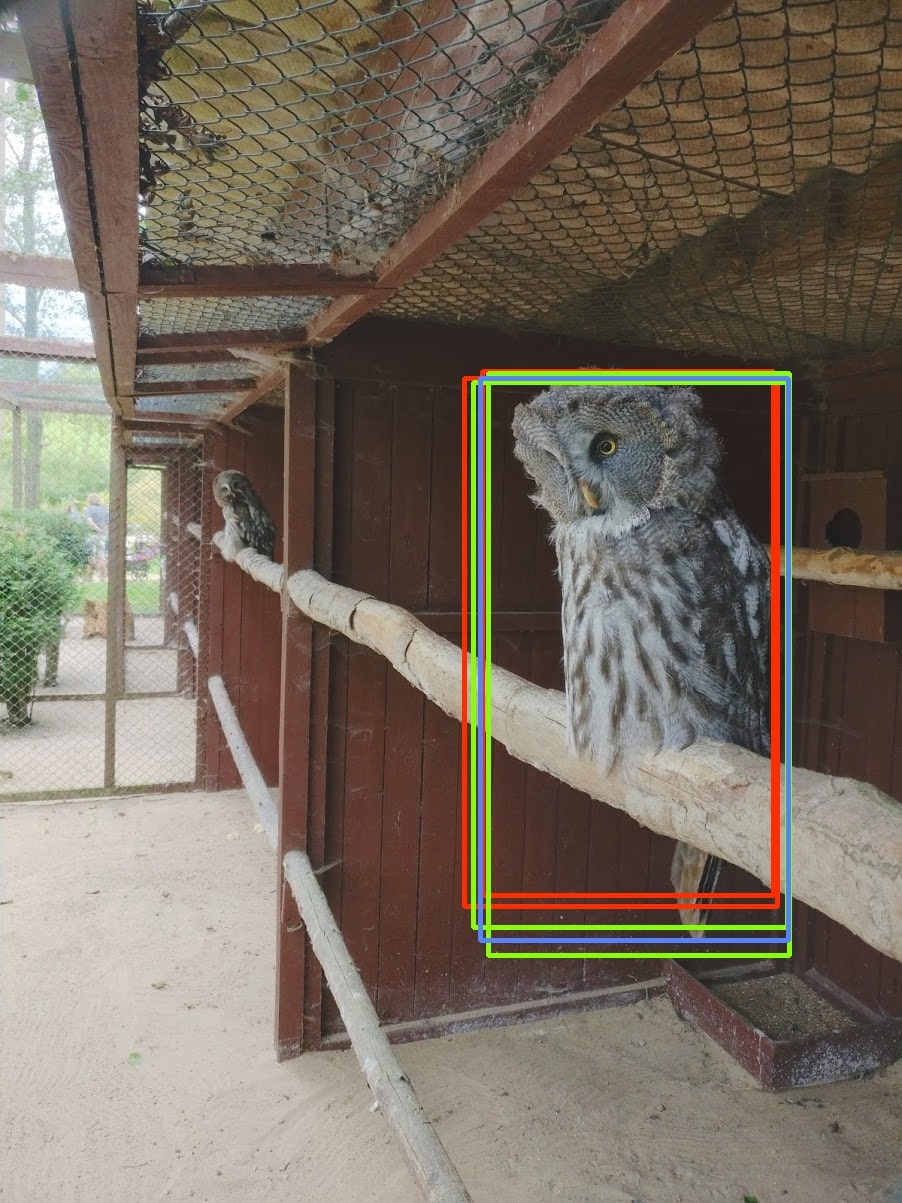}    &  \includegraphics[width=0.6\linewidth]{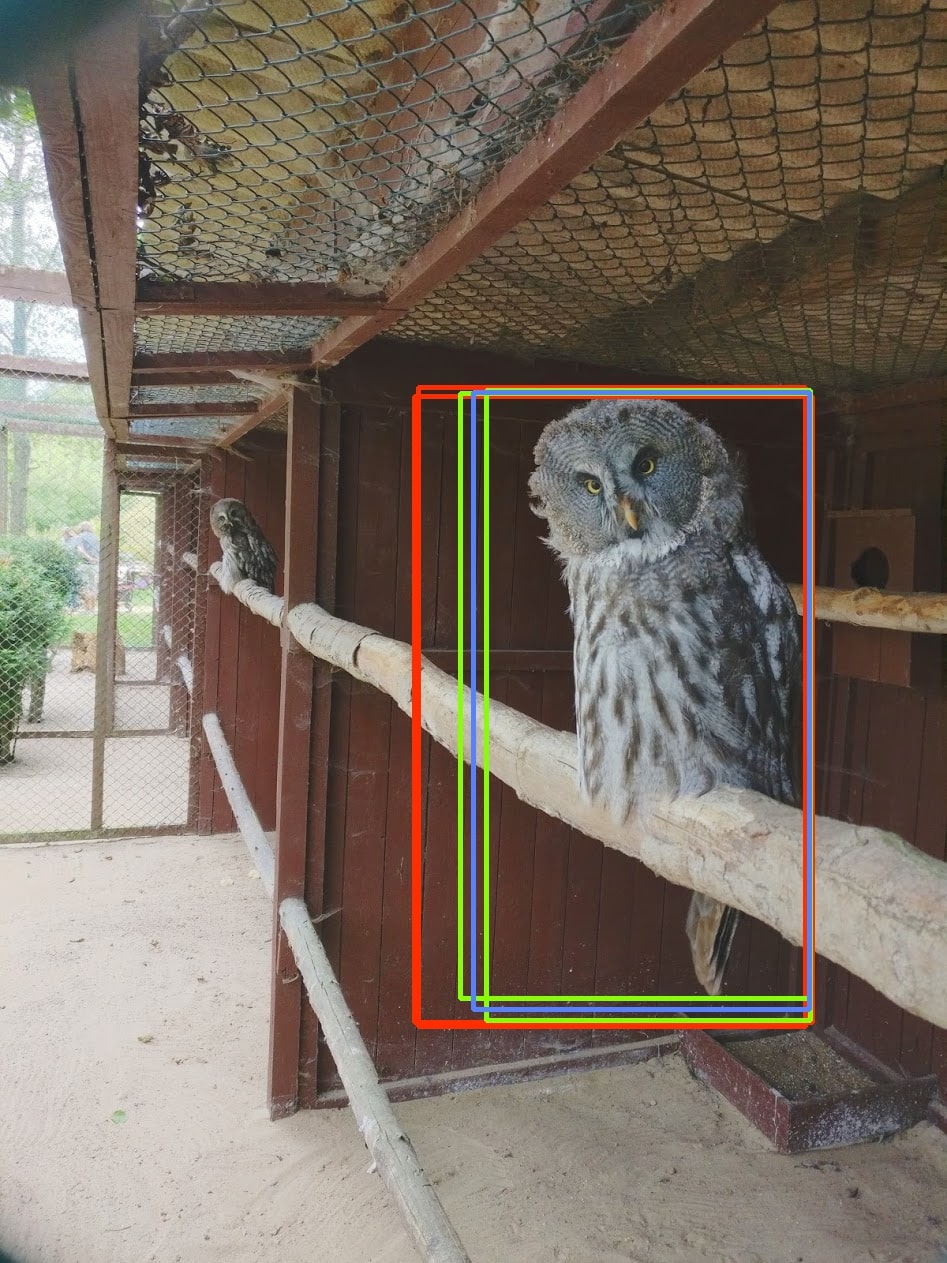}\\
      (e) Relative importance constraint & (f) Relative importance constraint \\
      Weights: (\textbf{0.46}, 0.18, 0.18, \textbf{0.18})   & Weights: (\textbf{0.26}, 0.24, \textbf{0.26}, 0.23) \\ 
      Sparse weights: (0.71, 0.29) &  Sparse weights: (0.50, 0.50) \\ 
    \end{tabular}
    }
    \caption{Comparison of bounding box predictions by distinct experts and the distribution of weights of \textit{2Block-MoE} with different load balancing  constraints for two views on the same object. Bounding box predictions of two active experts are green, and predictions of inactive experts are red. Raw image source: own recording of an author. }
    \label{fig:bbox_detectormoe_comparison_owl}
\end{figure}

Visual comparison of the bounding boxes (see
Figure~ \ref{fig:bbox_detectormoe_comparison_owl}) for a selected image reveals, that all experts have problems estimating the precise object boundaries for an atypical pose, whereas predictions on an object with a clear front view vary less. The gate is not always able to pick the single best expert. Still, the inactive experts would both overestimate the bottom of the bounding box, while the selected experts both predict the bottom tightly. Predictions made by distinct experts in the hard-constrained model vary less than for models trained with soft constraints.

To analyze the behavior of distinct experts in the regressor, we compute the averagely assigned weights to each regression expert during the evaluation and break down the weight assignments into the different feature map levels $P_3$ to $P_7$. On each feature map level, the gating network computes separate weights and consequently selects experts independently of other levels. We compute the average weight assignments on the sparse weight vectors after selecting the $k=2$ active experts (see Table \ref{tab:detectorMoE_regressor_avg_weighting}).  

\begin{table}[t]
	\centering
	\caption[Average weight assignment to distinct experts during evaluation]{Averagely assigned weights to each distinct \textit{2Block-MoE} \textbf{regression} expert during evaluation on the validation set 2017. Results are computed on the sparse vectors. Levels correspond to the feature map levels of RetinaNet. We highlight the top weights on each feature map level.}
	\label{tab:detectorMoE_regressor_avg_weighting}
\begin{tabular}{|r|lcccc|}
			\hline
			Constraint & Level       & Expert 1    & Expert 2     & Expert 3     & Expert 4 \\\hline
   
			& $P_3$      & \textbf{65.54\%} & 0.00\%           & 0.16\%   & 34.30\%          \\
			 KL-divergence & $P_4$      & \textbf{56.44\%} & 0.04\%           & 0.52\%   & 42.99\%          \\
			 loss & $P_5$      & 2.32\%           & 2.25\%           & 47.43\%  & \textbf{48.00\%} \\
			 & $P_6$      & 0.01\%           & \textbf{55.83\%} & 43.94\%  & 0.22\%           \\
   			& $P_7$      & 0.04\%           & \textbf{66.49\%} & 33.37\%  & 0.10\%           \\\hline

         & $P_3$      & 0.03\%           & 0.00\%           & 35.24\%  & \textbf{64.73\%} \\
			Importance & $P_4$      & 0.51\%           & 0.01\%           & 41.19\%  & \textbf{58.28\%} \\
			loss & $P_5$      & \textbf{50.10\%} & 1.60\%           & 47.07\%  & 1.23\%           \\
			 & $P_6$      & 47.34\%          & \textbf{52.59\%} & 0.07\%   & 0.00\%           \\
			& $P_7$      & 27.36\%          & \textbf{71.28\%} & 1.27\%   & 0.09\%           \\\hline
      
			   &$P_3$ & \textbf{67.75\%} & 2.62\%           & 13.85\%          & 15.79\%          \\
			Relative  &$P_4$ & 33.00\%          & 16.23\%          & \textbf{47.13\%} & 3.64\%           \\
			importance  & $P_5$ & 0.11\%           & \textbf{50.90\%} & 46.77\%          & 2.21\%           \\
			  constraint& $P_6$ & 9.69\%           & 29.81\%          & 23.81\%          & \textbf{36.69\%} \\
			 & $P_7$ & 0.29\%           & \textbf{54.30\%} & 44.03\%          & 1.38\%           \\
			\hline   
			   
		\end{tabular}
\end{table}

\begin{table}[t]
	\centering
	\caption[Average weight assignment to distinct  experts during evaluation]{Averagely assigned weights to each distinct \textit{2Block-MoE} \textbf{classification} expert during evaluation on the validation set 2017. Results are computed on the sparse vectors. Levels correspond to the feature map levels of RetinaNet. We highlight the top weights on each feature map level.}
	\label{tab:detectorMoE_classifier_avg_weighting}
		\begin{tabular}{|r|lcccc|}
			\hline
			Constraint & Level       & Expert 1    & Expert 2     & Expert 3     & Expert 4 \\\hline
			  & $P_3$ & 0.00\%           & 0.05\%           & \textbf{59.36\%} & 40.59\% \\
			KL-divergence & $P_4$ & 0.00\%           & 4.44\%           & \textbf{57.12\%} & 38.44\% \\
			 loss & $P_5$ & 1.42\%           & \textbf{58.42\%} & 8.15\%           & 32.01\% \\
			  & $P_6$ & 26.55\%          & \textbf{58.19\%} & 0.00\%           & 15.26\% \\
			& $P_7$ & \textbf{90.02\%} & 7.17\%           & 2.34\%           & 0.47\%  \\\hline

         & $P_3$ & 43.27\% & \textbf{56.70\%} & 0.03\%           & 0.00\%           \\
			Importance & $P_4$ & 39.70\% & \textbf{58.04\%} & 2.26\%           & 0.01\%           \\
			loss & $P_5$ & 31.15\% & 9.40\%           & \textbf{58.17\%} & 1.28\%           \\
			 & $P_6$ & 10.42\% & 0.00\%           & \textbf{58.58\%} & 31.00\%          \\
			& $P_7$ & 0.51\%  & 0.27\%           & 9.03\%           & \textbf{90.19\%} \\\hline
      
			  &$P_3$ & 0.06\%           & \textbf{69.68\%} & 11.67\% & 18.59\%          \\
			Relative & $P_4$ & 15.54\%          & 21.43\%          & 18.57\% & \textbf{44.46\%} \\
			importance&$P_5$ & \textbf{54.54\%} & 0.04\%           & 23.78\% & 21.64\%          \\
			 constraint&$P_6$ & 15.74\%          & 0.10\%           & 32.83\% & \textbf{51.33\%} \\
			& $P_7$ & \textbf{52.52\%} & 0.27\%           & 26.60\% & 20.62\%          \\
			\hline   
			   
		\end{tabular}
\end{table}

\begin{figure}[h]
\centering

\resizebox{1.0\linewidth}{!}{
    \begin{tabular}{c c}
     \includegraphics[width=0.6\linewidth]{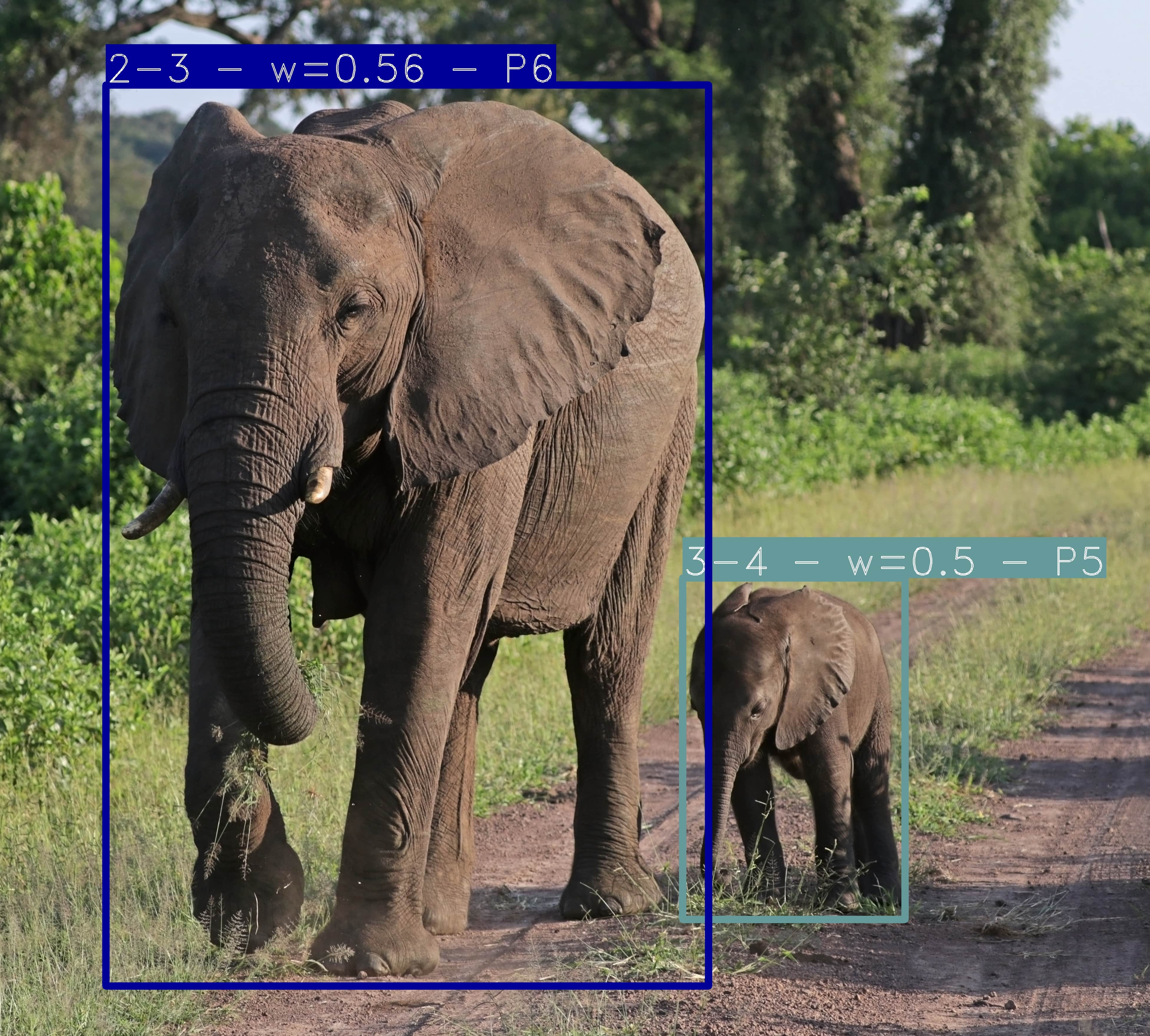}    &  \includegraphics[width=0.6\linewidth]{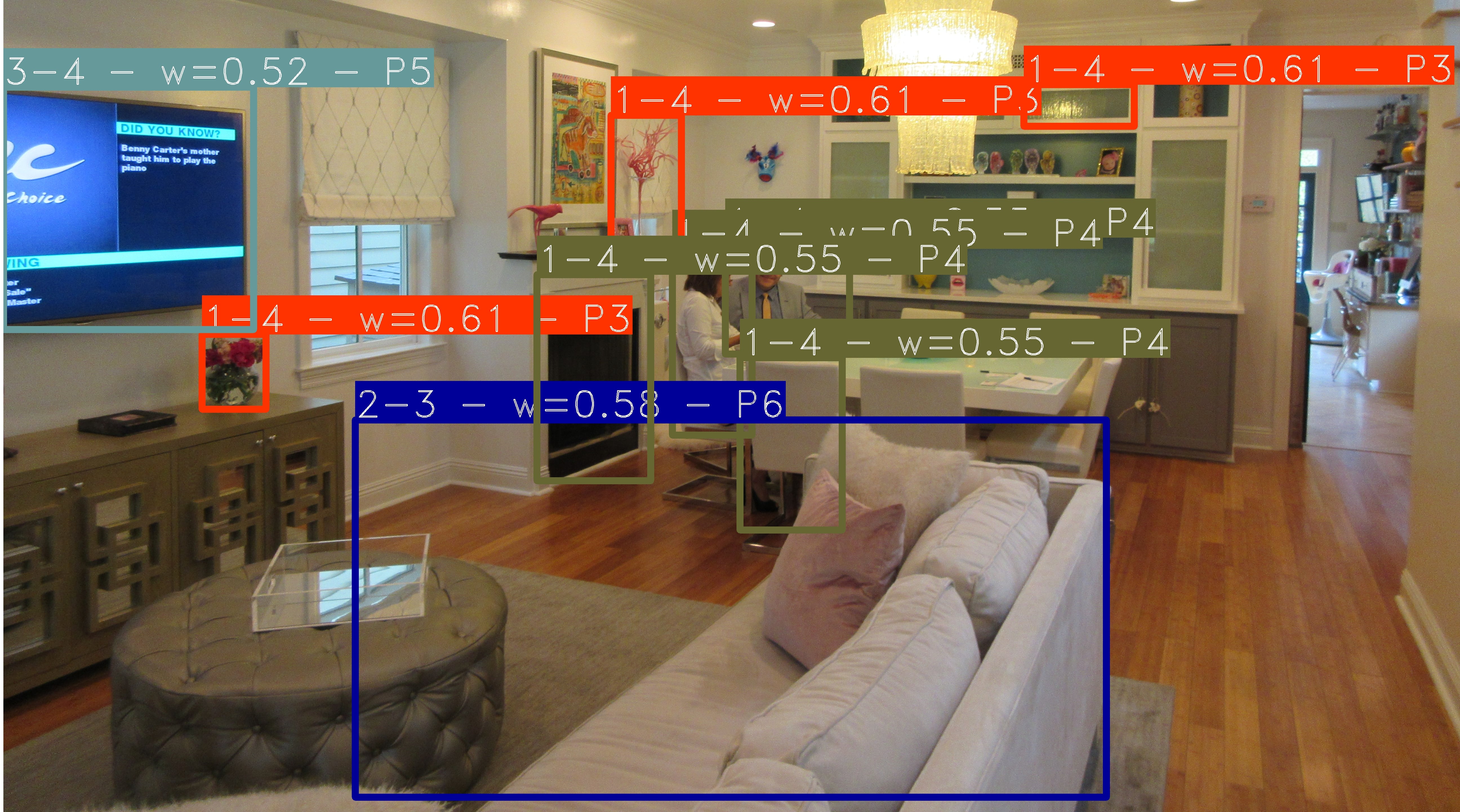}\\
      (a) Medium objects ($P_5$-$P_6$) & (b) Small to medium objects ($P_3$-$P_5$) \\
    \\
     \includegraphics[width=0.6\linewidth]{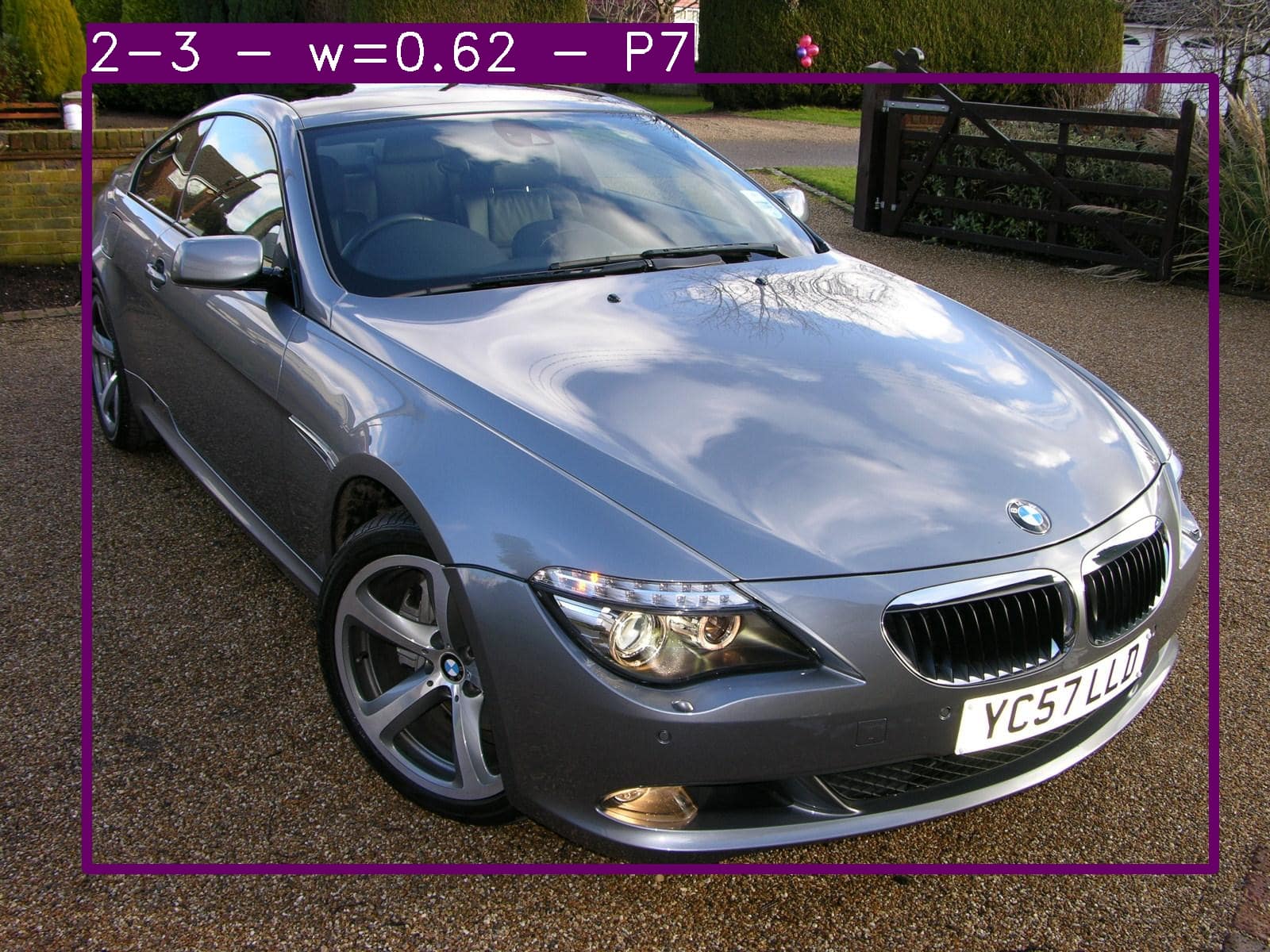}    &  \includegraphics[width=0.6\linewidth]{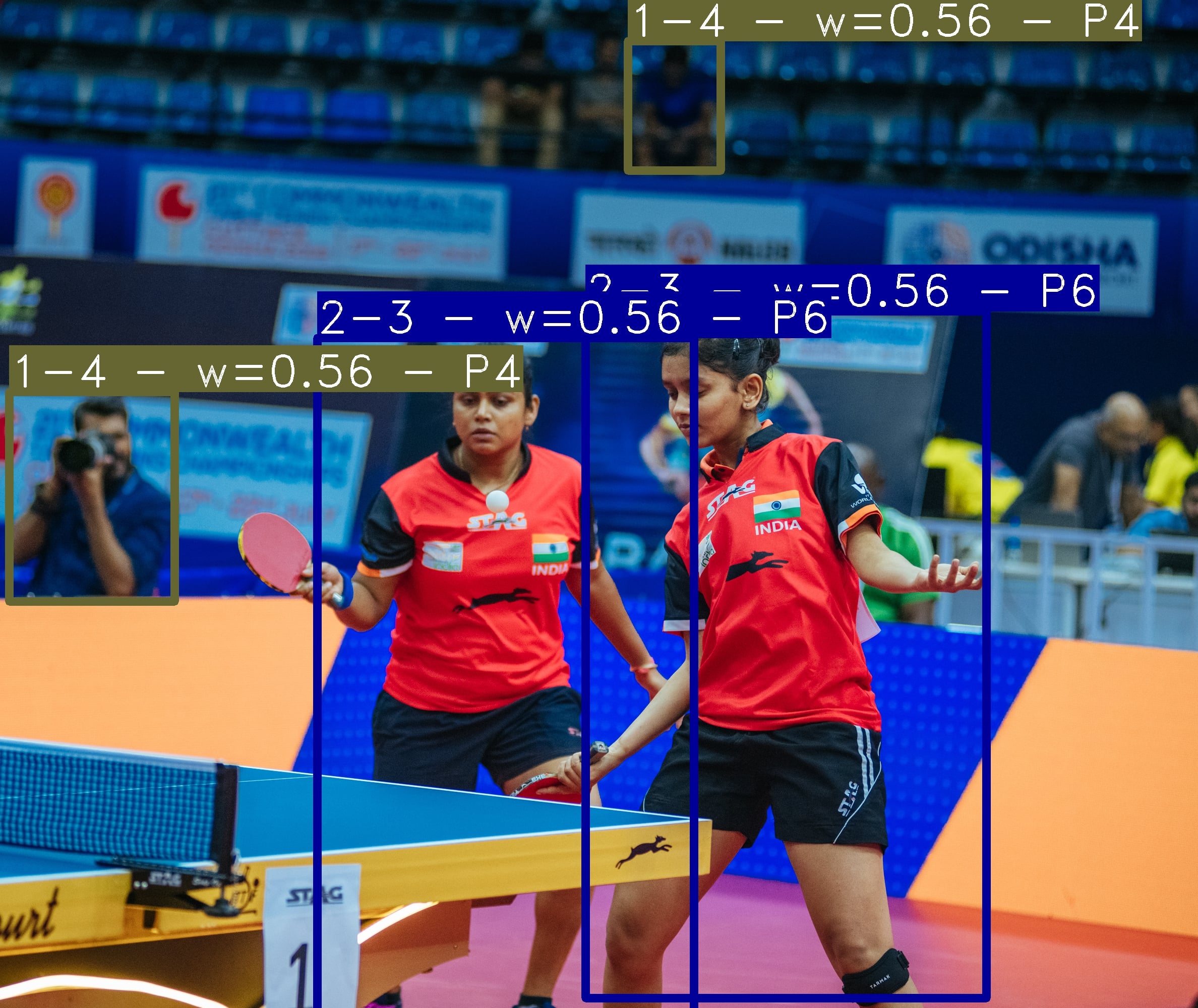}\\
      (c) Large object ($P_7$) & (d) Small to medium objects ($P_4$-$P_6$) \\
      \\

       \includegraphics[width=0.6\linewidth]{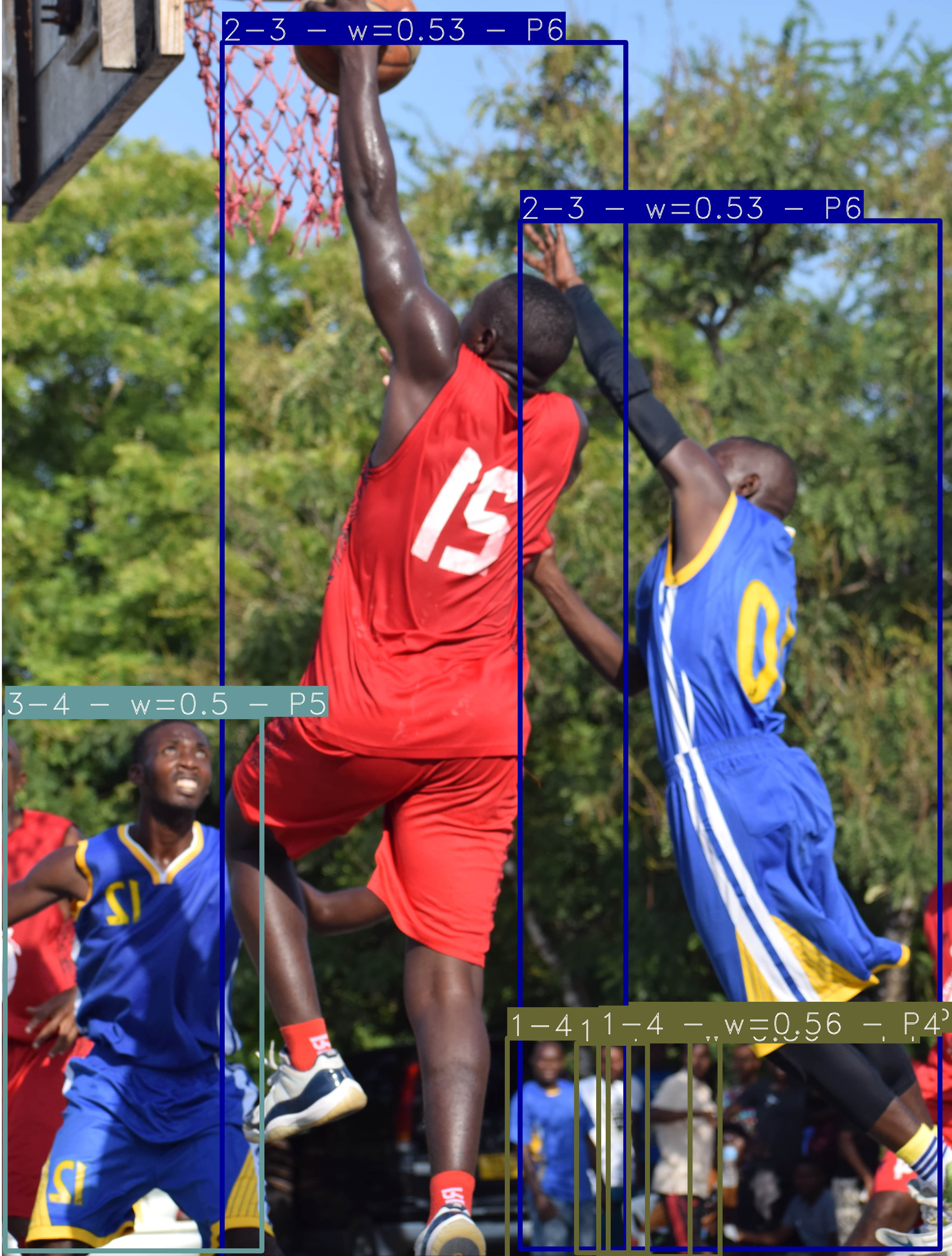}    &  \includegraphics[width=0.6\linewidth]{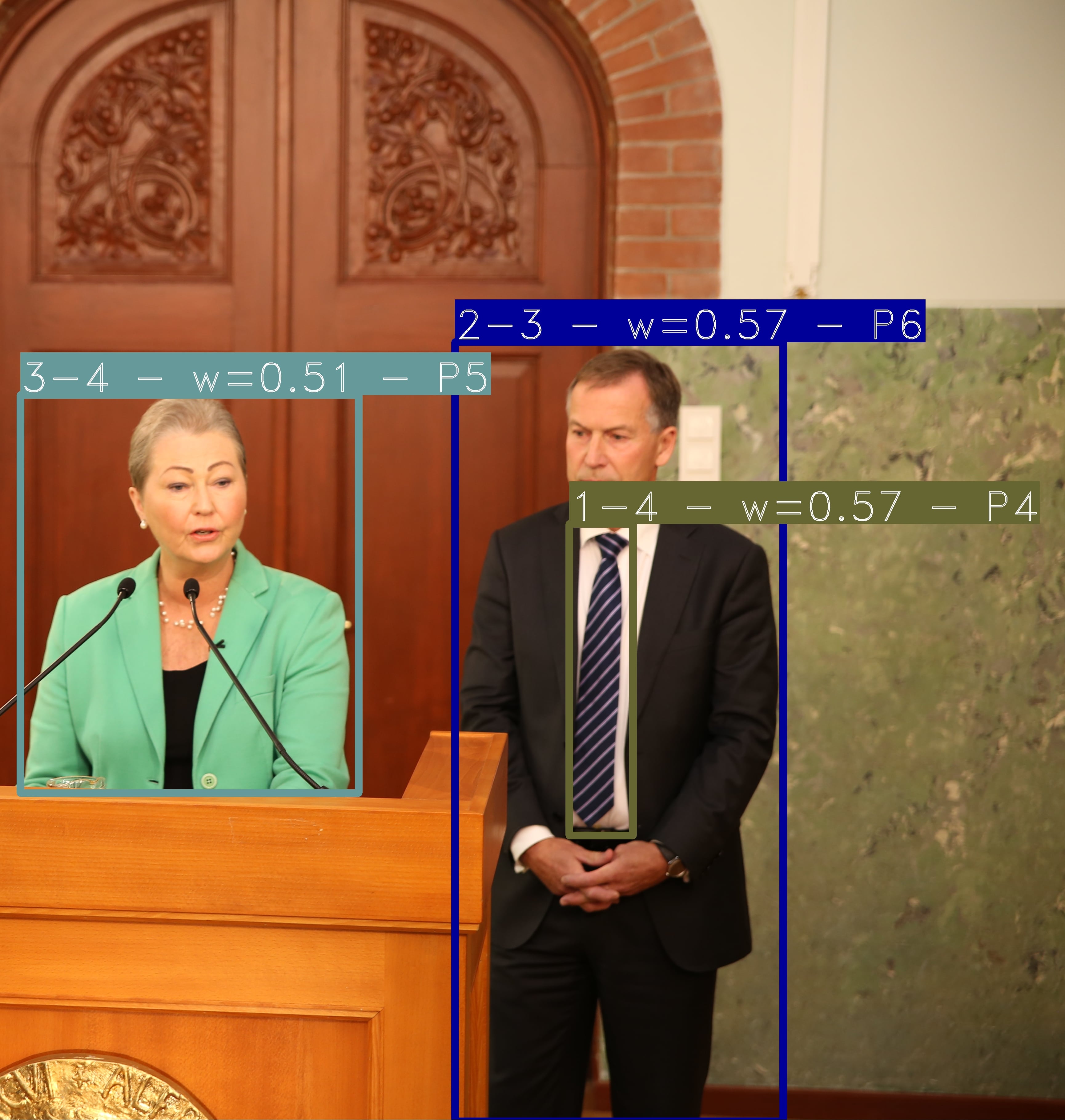}\\
      (e) Small to medium objects ($P_4$-$P_6$) & (f) Small to medium objects ($P_4$-$P_6$) \\
      
    \end{tabular}
    }
	\caption[BBox predictions on different feature map levels]{Bounding box predictions by \textit{2Block-MoE} with KL-divergence loss. Each bounding box caption includes the index of the two highest-weighted experts, the weight assigned to the top expert, and the feature map level on which the detection is made. Output images are partially cropped.}
     \label{fig:detectorMoE_KL_regression_per_level}
\end{figure}

We also illustrate the results in Figure \ref{fig:detectorMoE_KL_regression_per_level} by plotting the predicted bounding boxes for several images\footnote{Images taken from Wikimedia Commons, licensed under CC BY-SA 4.0 (b-d) and CC BY 4.0 (a,e,f). 
} with objects of varying scales. The results show that the gating network selects experts depending on the feature map levels. 

The hard-constrained gating network using the relative importance constraint assigns weights in a more diversified way (see Table \ref{tab:detectorMoE_regressor_avg_weighting}). Still, experts tend to be mainly utilized for different feature map levels. We conclude that distinct experts in the hard-constrained case perform better because they are activated on a larger number of different feature map levels and are trained to compute bounding boxes on different scales. Expert 1 is the only expert that is primarily used for detecting small objects. 

Altogether, we conclude that the gating network selects experts mainly depending on the input of different feature map levels. Even though we did not apply the constraints level-wise, the experts performed similarly accurate when detecting objects of different scales at different feature map levels. Improvements could be made by training specific experts on datasets, particularly for small or large objects. 

\textbf{Specialization of the \textit{2Block-MoE} in classifier}: we analyze the behavior of the classifier experts analogously by keeping $k_{reg}=2$ and assigning all weights in the classifier MoE to a specific expert (see Table~\ref{tab:detectorMoE_classifier_avg_weighting}). We observed that classifier experts vary to a greater extent than regressor experts. For the model with KL-divergence loss, Expert 1 stands out and performs significantly worse than the other experts. The expert is mostly utilized on feature map level $P_7$ with about 90\% of weight assignments. Consequently, the expert specializes in detecting large objects on this specific feature map size. 
Differences between the other experts in this model are still small but slightly larger than for the regressor experts. Experts in the hard-constrained model perform slightly better on their own and are closer to the MoE model utilizing $k_{cls}=2$ experts.

Weight assignments in the classifier MoE are similar to the regressor MoE, differences arise for levels $P_6$ and $P_7$.  
While the gating network in the regressor MoE focuses on two experts per level, the classifier's gating network utilizes three experts for $P_6$ and mostly a single expert for $P_7$. The hard-constrained model also assigns weights more diversified to its experts in the classifier MoE. 
\begin{figure}[t]
	\includegraphics[width=\linewidth]{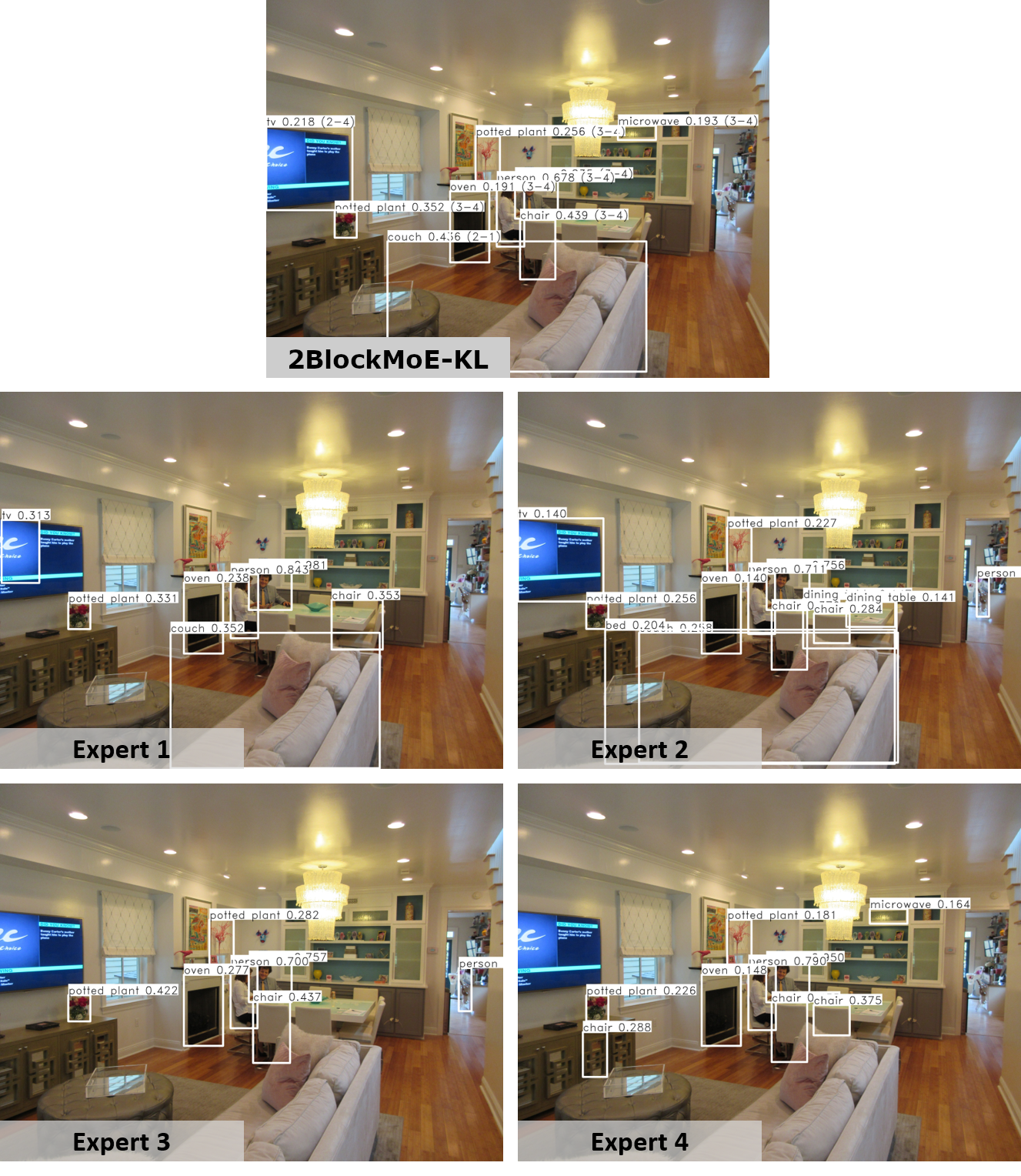}
	\caption{Comparison of different classifier expert predictions and the full MoE model. All detections are made by \textit{2BlockMoE} with KL-divergence loss. For each bounding box, we state the predicted class and the confidence score. We further state two active experts for each MoE prediction in brackets.}
    \label{fig:classifier}
\end{figure}

Figure \ref{fig:classifier} shows predictions of distinct classifier experts for an exemplary image. We set $k_{reg}=2$ and plot predictions using distinct classification experts and compare the experts' predictions to the full MoE model on the same input image. Note that all experts can detect relevant objects correctly but vary in their confidence. Experts 2 and 4 also falsely predict additional objects.

Overall, MoE layers embedded in the classifier and regressor subnets, the decision units in the network, allow us to gain insights into the decision processes of a model. We can separately analyze the predictions of each expert and compare them to the resulting MoE prediction. Our experiments have demonstrated, that distinct experts specialize in detecting objects of specific sizes.

\textbf{Impact of pre-training}: lastly, we investigated the behavior of MoE models using pre-trained expert weights. For this, we re-used baseline weights and added Gaussian noise to enforce expert specializations. We also trained \textit{Conv4} models by only replacing the $4^{th}$ convolutional layers in the regressor and classifier subnets with an MoE layer. For comparison, we also trained \textit{Conv4} models without pre-training.

Both models with pre-trained weights outperformed the baseline slightly and also outperformed models trained from scratch, whereas soft-constrained models showed larger improvements (see Table \ref{tab:DetectorMoE}).  The analysis of weight assignments (see Table \ref{tab:detectorMoE_pretrained_avg_weighting}) has shown again, that expert utilization is mainly dependent on the feature map levels. However, expert utilization is more scattered, and level boundaries are less rigid. We assume that for longer training durations, the gating network would refer stronger to one or two experts on each level, comparable to other MoE models. For the classifier MoE, assigned weightings are also  distributed better between different feature map levels

\begin{table}[t]
	\centering
	\caption[Average weight assignment to distinct experts during evaluation]{Averagely assigned weights to each distinct \textbf{regression} and \textbf{classification} expert of the \textit{2Block-MoE} with \textbf{pre-trained weights} using KL-divergence loss during evaluation on the validation set 2017. Levels correspond to the feature map levels of RetinaNet. We highlight the top weights on each feature map level.}
	\label{tab:detectorMoE_pretrained_avg_weighting}
		\begin{tabular}{|r|lcccc|}
			\hline
			  & Level       & Expert 1    & Expert 2     & Expert 3     & Expert 4 \\\hline
  		  & $P_3$ & 37.83\% & \textbf{48.78\%} & 0.64\%           & 12.76\% \\
			& $P_4$ & 36.75\% & \textbf{46.07\%} & 1.95\%           & 15.23\% \\
			Regressor & $P_5$ & 21.00\% & \textbf{38.26\%} & 15.33\%          & 25.40\% \\
			& $P_6$ & 8.46\%  & 24.40\%          & \textbf{38.56\%} & 28.58\% \\
			& $P_7$ & 3.20\%  & 30.58\%          & \textbf{50.67\%} & 15.55\% \\\hline

            & $P_3$ & 0.01\%           & 48.13\%          & \textbf{50.76\%} & 1.09\%  \\
			& $P_4$ & 0.75\%           & 41.80\%          & \textbf{46.31\%} & 11.14\% \\
			Classifier & $P_5$ & 29.08\%          & \textbf{29.53\%} & 22.19\%          & 19.20\% \\
			& $P_6$ & \textbf{54.14\%} & 22.64\%          & 8.07\%           & 15.15\% \\
			& $P_7$ & \textbf{55.03\%} & 18.59\%          & 9.42\%           & 16.95\% \\\hline
			   
		\end{tabular}
\end{table}

Overall, using pre-trained weights helps to increase performance compared to training from scratch, but obstructs expert specializations. Consequently, the decision-making process of a model becomes less transparent and less interpretable. This again stresses the existence of a trade-off between interpretability and expert specialization and model performance. 

\section{Conclusion}
In this work, we applied the sparsely-gated MoE layers to CNNs for computer vision tasks with the goal of increasing the model interpretability. We presented constraints to mitigate the dying expert problem, which tackle the issue from different angles and lead to different MoE behavior. Our analysis has revealed several interconnections between the proposed constraints on the one hand, and model performance as well as interpretability, on the other hand. Hard constraints result in better overall performance and generalized experts, although the mean importance constraint is particularly prone to the dying expert problem. Soft constraints, on the other side,  lead to better expert specialization. The usage of constraints thus helps to control the interplay between model performance, training stability, and expert specialization.

Our experiments have revealed inherent interpretability for two evaluated computer vision tasks. For the image classification task, experts focused on distinct repeating class groups, whereas, for object detection, they specialized in objects of distinct sizes. We hope that our insights pave the way for further research on the interpretability of deep neural networks.

\section*{Acknowledgement}

The research leading to these results is funded by the German Federal Ministry for Economic Affairs and Climate Action within the project “KI Absicherung“ (grant 19A19005W) and by KASTEL Security Research Labs. The authors would like to thank the consortium for the successful cooperation.

%%%%%%%%% REFERENCES
{\small
\bibliographystyle{IEEEtran}
\bibliography{literature}
}

\end{document}